\documentclass[journal]{IEEEtran}
\usepackage{graphicx}

\usepackage{color}
\usepackage[table]{xcolor}
\usepackage{colortbl}
\usepackage{bm}
\usepackage{amsmath}
\usepackage{epstopdf}

\usepackage{multicol}
\usepackage{multirow}
\usepackage{booktabs}
\usepackage{amssymb}

\usepackage{bbding}
\usepackage{stfloats}

\usepackage{lineno,hyperref}

\usepackage{makecell}
\usepackage{hhline}
\definecolor{linecolor}{rgb}{0.82, 0.94, 0.75}

\usepackage{booktabs} % For better table formatting

\definecolor{greenbg}{rgb}{0.9, 1.0, 0.9} % Define a light green background color

\newcommand{\tabincell}[2]{\begin{tabular}{@{}#1@{}}#2\end{tabular}}

\definecolor{lowred}{RGB}{238,18,137}

\definecolor{lowerred}{RGB}{255,110,180}

\newcommand{\dplus}[1]{\fontsize{6pt}{0.1em}\selectfont (\textbf{\textcolor{lowred}{#1}})}

\definecolor{rowunit}{RGB}{128,128,255}
\definecolor{evaunit01green}{RGB}{54,125,189}
\newcommand{\evagreen}[1]{\textcolor{evaunit01green}{#1}}
\newcommand{\dtplus}[1]{\fontsize{6pt}{0.1em}\selectfont (\textbf{\evagreen{#1}})}
\makeatletter
\renewcommand{\maketag@@@}[1]{\hbox{\m@th\normalsize\normalfont#1}}%
\makeatother

\usepackage{diagbox}

\ifCLASSINFOpdf
\else
\fi
\begin{document}
\title{Collaborative Low-Rank Adaptation for Pre-Trained Vision Transformers}

\author{Zheng~Liu,
        Jinchao Zhu,
        and~Gao~Huang,~\IEEEmembership{Member,~IEEE}% <-this % stops a space
\thanks{
Z.~Liu is with the School of Automation and Electrical Engineering, University of Science and Technology Beijing, Beijing 100083, China, and also with the Beijing Engineering Research Center of Industrial Spectrum Imaging, Beijing 100083, China.
J.~Zhu is with the College of Software, Nankai University, Tianjin 300350, China.
G.~Huang is with the Department of Automation, BNRist, Tsinghua University, Beijing 100084, China.
}% <-this % stops a space
\thanks{Corresponding author: J.~Zhu (e-mail: jczhu@nankai.edu.cn).}% <-this % stops a space
}


\markboth{Journal of \LaTeX\ Class Files,~Vol.~xx, No.~xx, December~2025}%
{Shell \MakeLowercase{\textit{et al.}}: Bare Demo of IEEEtran.cls for IEEE Journals}



\maketitle

\begin{abstract}
Low-rank adaptation (LoRA) has achieved remarkable success in fine-tuning pre-trained vision transformers for various downstream tasks. Existing studies mainly focus on exploring more parameter-efficient strategies or more effective representation learning schemes. However, these methods either sacrifice fine-tuning performance or introduce excessive trainable parameters, failing to strike a balance between learning performance and parameter efficiency. To address this problem, we propose a novel tuning method named collaborative low-rank adaptation (CLoRA) in this paper. CLoRA consists of base-space sharing and sample-agnostic diversity enhancement (SADE) components. To maintain parameter efficiency while expanding the learning capacity of low-rank modules (LRMs), base-space sharing allows all LRMs to share a set of down/up-projection spaces. In CLoRA, the low-rank matrices obtained from the shared spaces collaboratively construct each LRM. Since the representations extracted by these matrices may contain redundant information, SADE is employed to regularize the similarities among them to encourage diverse representations in the training process. We conduct extensive experiments on widely used image and point cloud datasets to evaluate the performance of CLoRA. Experimental results demonstrate that CLoRA strikes a better balance between learning performance and parameter efficiency, while requiring the fewest GFLOPs for point cloud analysis, compared with the state-of-the-art methods.
\end{abstract}


% \fi
\IEEEpeerreviewmaketitle

\section{Introduction}
\IEEEPARstart{V}{ision} transformers (ViTs) \cite{b1} have attained significant success across various computer vision tasks \cite{b2,b3,b4}. However, the strong performance relies on large amounts of training data. Consequently, a high-performing ViT is difficult to train from scratch in data-scarce scenarios \cite{b5}. To overcome the difficulty, a prevalent strategy is to first pre-train a ViT on large-scale datasets and subsequently fine-tune the entire model for target domains \cite{b6,b7}. Nevertheless, the strategy requires storing one entire ViT model for each downstream task, resulting in substantial storage costs.

Parameter-efficient fine-tuning (PEFT) \cite{b8,b9,b10} has become an active research area aimed at addressing the above challenge. Unlike full fine-tuning, PEFT methods, such as adapter \cite{b11}, prompt learning \cite{b12}, and low-rank adaptation (LoRA) \cite{b13}, normally introduce only a small number of trainable parameters while keeping the backbone model frozen. Among these methods, LoRA has attracted increasing attention from researchers and has been broadly discussed in recent years \cite{b14,b15,b16}.

\begin{figure}[!h]
\centering
\includegraphics[trim={8.8cm 3.15cm 0cm 4.5cm},clip,width=6.2in]{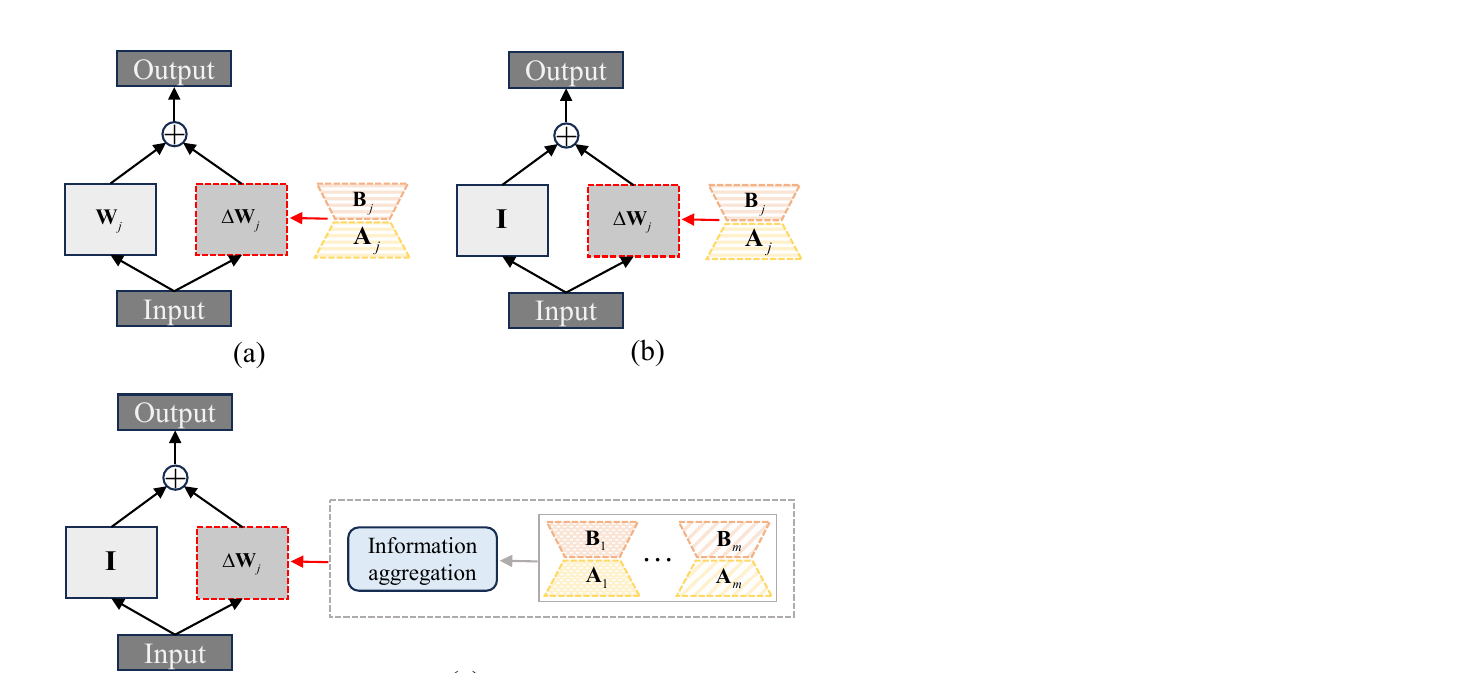}
\caption{Low-rank structures of (a) LoRA, (b) ARC, and (c) CLoRA.}
\label{fig1}
\end{figure}

Given a set of pre-trained weights $\{ {{\bf{W}}_i}\} _{i = 1}^m$ in a pre-trained backbone, where $m$ denotes the number of weights, LoRA assumes that their weight updates $\{ \Delta {{\bf{W}}_i}\} _{i = 1}^m$ are low-rank and can be learned by multiplying the corresponding low-rank down-projection matrices  $\{ {{\bf{A}}_i}\} _{i = 1}^m$ and up-projection matrices $\{ {{\bf{B}}_i}\} _{i = 1}^m$. During the fine-tuning process, LoRA adopts $m$ low-rank modules (LRMs) to capture the updates of $\{ {{\bf{W}}_i}\} _{i = 1}^m$. As  for the $j$-th LRM, its input is first projected into a low-dimensional space with dimension $r$ by down-projection matrix ${{\bf{A}}_j}$ and then is remapped back to the original space by up-projection matrix ${{\bf{B}}_j}$. After training, by substituting ${{\bf{W}}_j}$ with ${{\bf{W}}_j} + {{\bf{A}}_j}{{\bf{B}}_j}$, LoRA keeps the structural integrity of the original backbone. As numerous weights exist in a transformer-based backbone, applying LRMs to all of them would significantly increase the number of trainable parameters. Hence, LoRA typically employs LRMs to learn the updates of query and value projection weights. In ViT, each encoder layer primarily consists of a multi-head attention (MHA) block and a feed-forward network (FFN) block. However, LoRA may perform poorly on some downstream tasks because it only adapts a limited subset of the MHA blocks (query and value projection weights) while ignoring the FFN blocks. A related method named ARC \cite{b17} experimentally demonstrates that adapting the inputs to MHA and FFN blocks is more effective than updating the query and value projection weights, when tuning pre-trained ViT backbones to downstream tasks. Note that each tuned weight in ARC can be viewed as an identity matrix ${\bf{I}}$ with proper size. Taking the $j$-th LRM as an example, we show the tuning difference between LoRA and ARC in Fig.~\ref{fig1}.

As illustrated by existing LoRA-based studies \cite{b14,b15}, $r$ significantly affects the expressive power of LRMs, and it is hard to achieve satisfactory fine-tuning results with small $r$. There are two widely used approaches to solve this problem. The first one is directly enlarging the value of $r$ \cite{b18}. The second one is employing a group of low-rank matrices to learn the weight update in each inserted LRM \cite{b19}. However, both approaches introduce a large number of trainable parameters, which may lead to overfitting during fine-tuning. This raises the question: is it necessary to improve the fine-tuning performance by substantially increasing the number of trainable parameters? Upon closer examination, both methods essentially increase the rank upper bound of weight updates, which indicates that expanding the rank capacity is a feasible way to enhance the learning abilities of LRMs.

In LoRA, $m$ LRMs are independent of each other and the rank upper bound of each low-rank matrix is $r$. According to the subadditivity of matrix rank, we observe that the rank upper bound of $\sum\limits_{i = 1}^m {{{\bf{A}}_i}{{\bf{B}}_i}} $ is larger than $r$. This suggests that it is possible to enhance the learning capacity of each injected module without introducing extra parameters by exploiting the low-rank matrices contained in the others. Based on this observation, we present a simple but effective method termed as collaborative low-rank adaptation (CLoRA) for pre-trained ViTs. As shown in Fig.~\ref{fig1}, for the $j$-th LRM, CLoRA utilizes the information contained in all low-rank matrices $\{ {{\bf{A}}_i}{{\bf{B}}_i}\} _{i = 1}^m$ to collaboratively learn $\Delta {{\bf{W}}_j}$, whereas LoRA and ARC rely solely on the individual information in ${{\bf{A}}_j}{{\bf{B}}_j}$.

CLoRA consists of base-space sharing and sample-agnostic diversity enhancement (SADE) components. Under base-space sharing, $\{ \Delta {{\bf{W}}_i}\} _{i = 1}^m$ of all LRMs are collaboratively constructed by $\{ {{\bf{A}}_i}\} _{i = 1}^m$  and $\{ {{\bf{B}}_i}\} _{i = 1}^m$ which share the same base down-projection and up-projection spaces, respectively. Since each LRM in CLoRA combines multiple low-rank projection matrices, SADE is designed to enhance the diversity of the representations obtained by these projection matrices. The main contributions are as follows.

\begin{itemize}

 \item We present an effective base-space sharing scheme to combine the information contained in $\{ {{\bf{A}}_i}\} _{i = 1}^m$ and $\{ {{\bf{B}}_i}\} _{i = 1}^m$  during the learning process of each LRM, which significantly improves the learning capacity of LRMs and decreases the parameter count.

 \item To mitigate redundant information in the representations extracted from multiple low-rank matrices within each LRM, SADE efficiently regularizes the similarities among these representations during training.

 \item Extensive experiments on widely used image and point cloud datasets demonstrate the superiority of CLoRA over the state-of-the-art methods.
\end{itemize}

\section{Related work}

In recent years, parameter-efficient fine-tuning (PEFT) has emerged as a dominant approach for transferring the generalizable knowledge embedded in pre-trained transformer backbones to diverse downstream tasks. The mainstream studies can be roughly partitioned into five categories: prompt tuning, adapter tuning, reparameterization tuning, side-tuning, and hybrid tuning.

\textbf{Prompt tuning} inserts extra learnable prompts into pre-trained backbones to capture task-specific information \cite{b20}. For example, Jia et al. \cite{b12} prepend a set of trainable tokens to either the first encoder layer only or all encoder layers; Pei et al. \cite{b21} learn two-dimensional prompt maps with sizes matching the corresponding image token maps for encoder layers; Hu et al. \cite{b22} and Zhang et al. \cite{b23} aggregate the intermediate information of pre-trained backbones by adding additional tokens for the queries in MHA blocks; Nie et al. \cite{b24} first divide the pre-trained backbone into multiple stages and then construct stage-wise prompt blocks to generate task-specific discriminative prompts for downstream task inputs. \textbf{Adapter tuning} inserts trainable bottleneck modules into pre-trained backbones, where each module comprises a down-projection layer, a non-linear activation, and an up-projection layer \cite{b25,b26}. Adapter-based methods typically extract downstream task-specific representations by applying these modules to 1) learn the supplementary information for FFN blocks \cite{b27,b28} or additional transformer components \cite{b29} and 2) transform the input or output representations of core building blocks such as MHA and FFN blocks \cite{b30,b31}.  \textbf{Side-tuning} trains a separate lightweight side network in parallel with the frozen pre-trained backbone and fuses their outputs for final predictions \cite{b32}. Such side networks either integrate the features extracted from the pre-trained backbone sequentially \cite{b33,b34} or fuse the extracted features through parallel aggregation \cite{b35,b36}. Unlike the above three categories which usually bring extra inference costs, \textbf{reparameterization tuning} aims to avoid the inference overhead by tuning the parameters that are either part of the backbone or can be merged into the backbone weights after training \cite{b37}. Methods belonging to this category design some strategies to identify task-specific parameters \cite{b38} or apply the modules with linear structures that can be incorporated into existing layers to learn task-specific representations \cite{b39,b40}. \textbf{Hybrid tuning} combines methods from different categories. For example, Zhou et al. \cite{b41} inject prompt tokens and dynamic adapters into encoder layers; Zhang et al. \cite{b42} explore the optimal combination of adapters, learnable tokens and linear LRMs by neural architecture search techniques.

Owing to its ability to avoid extra computational overhead, reparameterization tuning has received growing interest in the deep learning community. As a representative method, LoRA \cite{b13} has been widely studied. LoRA assumes that the weight updates for downstream tasks can be captured by linear LRMs where each of them consists of a down-projection matrix (mapping inputs into a low space with dimension $r$) and an up-projection matrix (remapping subspace embeddings back to the original space). Most existing studies focus on exploring effective strategies to learn representative down-projection and up-projection matrices. For example, Liu et al. \cite{b14} divide pre-trained weights into magnitude and direction components and LRMs are employed to learn directional updates; Hayou et al. \cite{b43} assign different learning rates for down-projection and up-projection matrices; Lin et al. \cite{b44} learn projection matrices based on the singular value decomposition of pre-trained weights; Yang et al. \cite{b45} employ two vectors to dynamically generate down-projection and up-projection matrices. When the pre-trained weights for LRMs are set as identity matrices, these modules are normally adopted to adapt the inputs of some components (such as MHA and FFN blocks) contained in pre-trained backbones \cite{b17,b46}. In practice, $r$ has a significant influence on the performance of LoRA-based methods. A small $r$ usually results in poor learning abilities \cite{b14,b15} while a large $r$ greatly increases the number of trainable parameters. Currently, motivated by the success of mixture of experts (MoE) \cite{b47}, some researchers have begun to design strategies to combine multiple experts in LRM where each expert is composed of a pair of down-projection and up-projection matrices. MOLE \cite{b48} employs learnable gating functions to compose multiple experts. In HydraLoRA \cite{b49}, all experts in each LRM share the same down-projection matrix. LoRAMoE \cite{b50} assigns multiple experts to learn task-specific information for FFN blocks. However, these methods introduce extra inference overhead due to the nonlinear routers used to evaluate the contribution of experts. To preserve the original property of LoRA, some studies \cite{b19,b50} treat different experts equally.

Enlarging the value of $r$ and combining multiple experts (a pair of down-projection and up-projection matrices) in LRMs are two effective ways to enhance the fine-tuning performance of LoRA-based methods. Both ways expand the rank upper bound of learned low-rank matrices, indicating that increasing the rank upper bound is a feasible approach to improve the learning abilities of these LRMs. Unfortunately, they bring a large number of trainable parameters, resulting in a parameter-inefficiency problem. Our proposed method (CLoRA) also incorporates multiple experts, but it is much more parameter-efficient than existing LoRA-based ensemble methods due to the designed base-space sharing scheme. Besides, during the training process, CLoRA encourages learning diverse experts by regularizing the similarities among the representations extracted by these experts, which is usually ignored by existing methods.

\section{Preliminaries}
\subsection{Vision Transformer}

The architecture of a standard vision transformer (ViT) \cite{b1} is composed of a patch embedding layer, $L$ stacked encoder layers, and a prediction head.

In ViT, the patch embedding layer is employed to transform input images into suitable embeddings for the subsequent layers. Specifically, given an input image ${\bf{X}} \in {R^{h \times w \times 3}}$, it first converts ${\bf{X}}$ into a sequence of $n$ flattened patches $[{{\bf{x}}_1}; \cdots ;{{\bf{x}}_n}] \in {R^{n \times 3{m^2}}}$ and then obtains embedding tokens $[{{\bf{x}}_1}{\bf{W}}; \cdots ;{{\bf{x}}_n}{\bf{W}}] \in {R^{n \times d}}$  with mapping matrix ${\bf{W}} \in {R^{3{m^2} \times d}}$,  where $(h,w)$ and $(m,m)$ are the resolutions of ${\bf{X}}$ and its patches, $d$ is the dimension of each token, and $n = \frac{{hw}}{{{m^2}}}$. The mathematical expression of this layer is formulated as
\begin{equation}
{{\bf{Z}}^0} = [{{\bf{x}}_{class}};{{\bf{x}}_1}{\bf{W}}; \cdots ;{{\bf{x}}_n}{\bf{W}}] + {{\bf{E}}_{pos}}, \label{eq1}
\end{equation}
where ${{\bf{x}}_{class}} \in {R^{1 \times d}}$ is the prepended class token, ${{\bf{E}}_{pos}} \in {R^{(n + 1) \times d}}$ denotes the position embeddings of tokens and ${{\bf{Z}}^0}$ represents the transformed token set of this layer.

Encoder layers inserted after the patch embedding layer are designed to process the output information ${{\bf{Z}}^0}$. Each encoder layer consists of a MHA block and a FFN block, where each of them is preceded by a layer normalization (LN) module. As for the $l$-th ($1 \le l \le L$) encoder layer, its mathematical expression is
\begin{equation}
{{\bf{\tilde Z}}^l} = {\rm{MHA}}({\rm{LN}}({{\bf{Z}}^{l - 1}})) + {{\bf{Z}}^{l - 1}},
\label{eq2}
\end{equation}
\begin{equation}
{{\bf{Z}}^l} = {\rm{FFN}}({\rm{LN}}({{\bf{\tilde Z}}^l})) + {{\bf{\tilde Z}}^l},
\label{eq3}
\end{equation}
where ${{\bf{\tilde Z}}^l}$  and ${{\bf{Z}}^l}$ denote the outputs of the MHA and FFN blocks. Note that ${{\bf{Z}}^l}$ is also the output of the encoder layer.

Through equations (1)-(3), we can calculate the class token's representation at the $L$-th encoder layer. Let ${\bf{z}}_{class}^L$ denote the representation of ${{\bf{x}}_{class}}$ contained in ${{\bf{Z}}^L}$, the prediction result ${\bf{y}}$ for ${\bf{X}}$ is
\begin{equation}
{\bf{y}} = {\rm{PH}}({\rm{LN}}({\bf{z}}_{class}^L)),
\label{eq4}
\end{equation}
where ${\rm{PH}}( \cdot )$ represents a lightweight classifier.

\subsection{Low-Rank Adaptation}

LoRA \cite{b13} is based on the hypothesis that weight updates lie in low-rank spaces when adapting a pre-trained backbone to downstream tasks.

When fine-tuning a ViT backbone with LoRA, the updated weights are typically $d \times d$ matrices. Therefore, for the $j$-th LRM, we have ${{\bf{W}}_j} \in {R^{d \times d}}$ and $\Delta {{\bf{W}}_j} \in {R^{d \times d}}$. As shown in Fig.~\ref{fig1}(a), $\Delta {{\bf{W}}_j}$ is parameterized as ${{\bf{A}}_j}{{\bf{B}}_j}$ to ensure the low-rank property, where ${{\bf{A}}_j} \in {R^{d \times r}}$ and ${{\bf{B}}_j} \in {R^{r \times d}}$ are down-projection and up-projection matrices. In the $j$-th LRM, the output representation ${\bf{X}}_{output}^j$ is obtained by
\begin{equation}
{\bf{X}}_{output}^j = {\bf{X}}_{input}^j({{\bf{W}}_j} + \Delta {{\bf{W}}_j}) = {\bf{X}}_{input}^j({{\bf{W}}_j} + {{\bf{A}}_j}{{\bf{B}}_j}),
\label{eq5}
\end{equation}
where ${\bf{X}}_{input}^j$ is the input of this module.

During the fine-tuning process, LoRA freezes the weights $\{ {{\bf{W}}_i}\} _{i = 1}^m$ and only trains $\{ {{\bf{A}}_i}{{\bf{B}}_i}\} _{i = 1}^m$ with the data from downstream tasks. Compared with full fine-tuning, LoRA greatly reduces the number of trainable parameters. At the inference stage,   $\{ {{\bf{A}}_i}{{\bf{B}}_i}\} _{i = 1}^m$ can be directly merged into the corresponding weights through replacing $\{ {{\bf{W}}_i}\} _{i = 1}^m$ with $\{ {{\bf{W}}_i} + {{\bf{A}}_i}{{\bf{B}}_i}\} _{i = 1}^m$, avoiding additional computational and storage costs.

\begin{figure*}[!htb]
\centering
\includegraphics[trim={2.0cm 5.0cm -11.1cm 2.0cm},clip,width=10.5in]{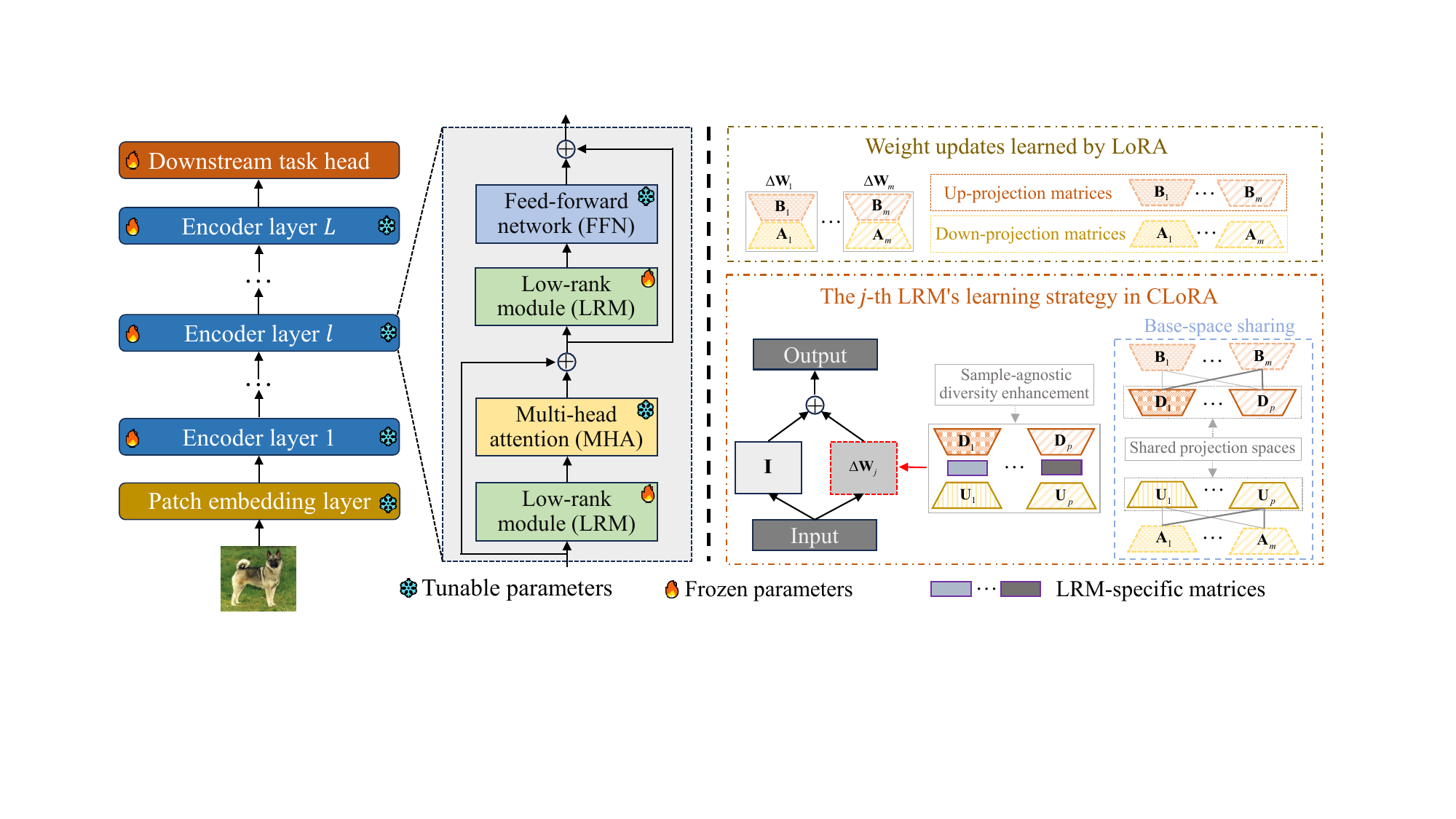}
\caption{Illustration of CLoRA. In LoRA, LRMs are applied to adapt the input representations of MHA and FFN blocks, yielding $\{ \Delta {{\bf{W}}_i} = {{\bf{A}}_i}{{\bf{B}}_i}\} _{i = 1}^m$, where $\{ {{\bf{A}}_i}{{\bf{B}}_i}\} _{i = 1}^m$ are independent of each other. By contrast, each LRM in CLoRA effectively and efficiently utilizes the information contained in $\{ {{\bf{A}}_i}{{\bf{B}}_i}\} _{i = 1}^m$  through the base-space sharing and SADE components.}
\label{fig2}
\end{figure*}

\section{Collaborative low-rank adaptation}

Inspired by the work of LoRA \cite{b13} and ARC \cite{b17}, we show the architecture of CLoRA in Fig.~\ref{fig2}. As displayed, two low-rank modules (LRMs) are employed to adapt the input representations of the MHA and FFN blocks in each encoder layer, respectively. To reduce the number of trainable parameters and simultaneously increase the rank upper bounds of $\{ \Delta {{\bf{W}}_i}\} _{i = 1}^m$, we introduce a novel base-space sharing scheme where $\{ \Delta {{\bf{W}}_i}\} _{i = 1}^m$  are collaboratively parameterized by down/up-projection matrices ($\{ {{\bf{A}}_i}\} _{i = 1}^m$/$\{ {{\bf{B}}_i}\} _{i = 1}^m$) which share common down/up-projection spaces. Specifically, each of $\{ {{\bf{A}}_i}\} _{i = 1}^m$ (or $\{ {{\bf{B}}_i}\} _{i = 1}^m$) can be linearly represented by shared down-projection matrices $\{ {{\bf{D}}_h}\} _{h = 1}^p$ (or up-projection matrices $\{ {{\bf{U}}_h}\} _{h = 1}^p$), where $p < m$. Under this scheme, $\{ \Delta {{\bf{W}}_i}\} _{i = 1}^m$ are learned via LRM-specific matrices and the shared down/up-projection matrices. As a result, each of $\{ \Delta {{\bf{W}}_i}\} _{i = 1}^m$ is the combination of multiple low-rank projection matrices. Finally, sample-agnostic diversity enhancement (SADE) is designed to improve the quality of $\{ \Delta {{\bf{W}}_i}\} _{i = 1}^m$ by regularizing the similarities among the representations extracted by these matrices during training.

\subsection{Base-Space Sharing}

When LRMs in LoRA are employed to tune the input representations of MHA and FFN blocks, as in ARC, the tuned weights in pre-trained backbones can be viewed as identity matrices. Correspondingly, for the $j$-th LRM, its output ${\bf{X}}_{output}^j$ can be obtained by
\begin{equation}
{\bf{X}}_{output}^j = {\bf{X}}_{input}^j({\bf{I}} + \Delta {{\bf{W}}_j}) = {\bf{X}}_{input}^j + {\bf{X}}_{input}^j{{\bf{A}}_j}{{\bf{B}}_j}.
\label{eq6}
\end{equation}

The above equation indicates that the transformed representation ${\bf{X}}_{output}^j$ is the sum of the original representation and its update. In this case, the update of ${\bf{X}}_{input}^j$ is learned by low-rank matrix ${{\bf{A}}_j}{{\bf{B}}_j}$. Since $r \ll d$, we have $rank({{\bf{A}}_j}{{\bf{B}}_j}) \le r$ where $rank( \cdot )$ represents the rank of ${{\bf{A}}_j}{{\bf{B}}_j}$. $r$ is an essential hyper-parameter for LoRA. In practice, a small $r$ usually limits the fine-tuning abilities on pre-trained models, while a large $r$ can bring numerous trainable parameters, resulting in parameter inefficiency. Actually, $r$ affects the fine-tuning performance by controlling the upper bound of $\Delta {{\bf{W}}_j}$'s rank. A small (or large) $r$ means a low (or high) upper bound on the rank. If $r$ is set as a small value, the learned low-rank space may have difficulty approximating the actual low-rank space, because the rank upper bound is limited. Therefore, enlarging the upper bound of rank is a feasible way to enhance the fine-tuning performance when $r$ is small.

In LoRA-based tuning methods, low-rank matrices $\{ {{\bf{A}}_i}{{\bf{B}}_i}\} _{i = 1}^m$ inserted into pre-trained backbones are applied to extract different representations and $\{ {{\bf{A}}_i}{{\bf{B}}_i}\} _{i = 1}^m$  are normally different. Since $rank(\sum\limits_{i = 1}^m {{{\bf{A}}_i}{{\bf{B}}_i}} ) \le \sum\limits_{i = 1}^m {rank} ({{\bf{A}}_i}{{\bf{B}}_i}) \le mr$, if we properly utilize all the inserted low-rank matrices, the rank upper bound can be increased to $mr$, thus improving the fine-tuning abilities. A simple idea is to learn each representation update by $\{ {{\bf{A}}_i}{{\bf{B}}_i}\} _{i = 1}^m$. For example, in the $j$-th LRM, the output ${\bf{X}}_{output}^j$ is formulated as
\begin{equation}
{\bf{X}}_{output}^j = {\bf{X}}_{input}^j({\bf{I}} + \Delta {{\bf{W}}_j}) = {\bf{X}}_{input}^j + {\bf{X}}_{input}^j\sum\limits_{i = 1}^m {{{\bf{A}}_i}{{\bf{B}}_i}} .
\label{eq7}
\end{equation}

For the $m$ LRMs at different locations, their input representation updates usually have different requirements for $\{ {{\bf{A}}_i}{{\bf{B}}_i}\} _{i = 1}^m$, indicating that some optimization conflicts may exist in the whole tuning process. To avoid the potential problem, we design LRM-specific transformation matrices $\{ {{\bf{\Lambda }}_{i,j}}\} _{i = 1}^m$ for $\Delta {{\bf{W}}_j}$ and then reformulate equation (7) as
\begin{equation}
{\bf{X}}_{output}^j = {\bf{X}}_{input}^j + {\bf{X}}_{input}^j\sum\limits_{i = 1}^m {{{\bf{A}}_i}{{\bf{\Lambda }}_{i,j}}{{\bf{B}}_i}} ,
\label{eq8}
\end{equation}
where ${{\bf{\Lambda }}_{ij}} \in {R^{r \times r}}$ is $\Delta {{\bf{W}}_j}$'s specific transformation matrix used to utilize the information contained in ${{\bf{A}}_i}{{\bf{B}}_i}$.  Note that ${{\bf{\Lambda }}_{ij}} = {\bf{I}}$ if $j = i$.

However, equation (8) will introduce more trainable parameters than those of LoRA. To lower the introduced parameter count, we assume that $\{ {{\bf{A}}_i}{{\bf{B}}_i}\} _{i = 1}^m$  share the same low-rank spaces $\{ {S_h}\} _{h = 1}^p$ ($p < m$). Under this assumption, each of $\{ {{\bf{A}}_i}{{\bf{B}}_i}\} _{i = 1}^m$ can be divided into multiple component matrices in spaces $\{ {S_h}\} _{h = 1}^p$. We denote the component matrices for ${{\bf{A}}_i}$  and ${{\bf{B}}_i}$ in the  $h$-th low-rank space ${S_h}$ as ${\bf{A}}_i^h$  and ${\bf{B}}_i^h$. Suppose space ${S_h}$ can be represented by ${{\bf{D}}_h}{{\bf{U}}_h}$, where ${{\bf{D}}_h}$ and ${{\bf{U}}_h}$ denote the base down-projection and up-projection matrices, respectively. Assuming matrices ${\bf{T}}_i^h \in {R^{r \times r}}$ and ${\bf{R}}_i^h \in {R^{r \times r}}$ are the representations for ${\bf{A}}_i^h$ and ${\bf{B}}_i^h$ under base projection matrices ${{\bf{D}}_h}$ and ${{\bf{U}}_h}$, we have ${\bf{A}}_i^h = {{\bf{D}}_h}{\bf{T}}_i^h$ and ${\bf{B}}_i^h = {\bf{R}}_i^h{{\bf{U}}_h}$. Accordingly, we have ${{\bf{A}}_i} = \sum\limits_{h = 1}^p {{\bf{A}}_i^h}  = \sum\limits_{h = 1}^p {{{\bf{D}}_h}{\bf{T}}_i^h} $ and ${{\bf{B}}_i} = \sum\limits_{h = 1}^p {{\bf{B}}_i^h}  = \sum\limits_{h = 1}^p {{\bf{R}}_i^h{{\bf{U}}_h}} $. Similarly, we also divide $\Delta {{\bf{W}}_j}$ into multiple component matrices in spaces $\{ {S_h}\} _{h = 1}^p$. For $\Delta {{\bf{W}}_j}$, its component matrix $\Delta {\bf{W}}_j^h$ in space ${S_h}$ is obtained as follows.
\begin{equation}
\Delta {\bf{W}}_j^h = \sum\limits_{i = 1}^m {{\bf{A}}_i^h{\bf{\Lambda }}_{i,j}^h{\bf{B}}_i^h}  = \sum\limits_{i = 1}^m {{{\bf{D}}_h}{\bf{T}}_i^h{\bf{\Lambda }}_{i,j}^h{\bf{R}}_i^h{{\bf{U}}_h},}
\label{eq9}
\end{equation}
where ${\bf{\Lambda }}_{i,j}^h$ is the corresponding component of ${{\bf{\Lambda }}_{i,j}}$ in space ${S_h}$.

Based on equation (9), we have
\begin{equation}
\Delta {{\bf{W}}_j} = \sum\limits_{h = 1}^p {\Delta {\bf{W}}_j^h}  = \sum\limits_{h = 1}^p {{{\bf{D}}_h}{\bf{Q}}_i^h} {{\bf{U}}_h},
\label{eq10}
\end{equation}
where ${\bf{Q}}_h^j = \sum\limits_{i = 1}^m {{\bf{T}}_i^h{\bf{\Lambda }}_{i,j}^h{\bf{R}}_i^h} $.

As analyzed above, the output of the $j$-th LRM in CLoRA is formulated as
\begin{equation}
{\bf{X}}_{output}^j = {\bf{X}}_{input}^j + {\bf{X}}_{input}^j\sum\limits_{h = 1}^p {{{\bf{D}}_h}{\bf{Q}}_h^j} {{\bf{U}}_h}.
\label{eq11}
\end{equation}
All components in ${\bf{Q}}_h^j$ are trainable during the fine-tuning process, which implies ${\bf{Q}}_h^j$ is also trainable. In equation (11), we can directly train a suitable ${\bf{Q}}_h^j$ using the data of downstream tasks and thus ignore the learning processes of its components.

As shown in Fig.~\ref{fig2}, two LRMs are inserted before the MHA and FFN blocks of each encoder layer. Taking the $l$-th encoder layer as an example, the ($2l-1$)-th and  $2l$-th LRMs are inserted in this layer. Let ${\bf{\tilde Z}}_{MHA}^l$ and ${\bf{Z}}_{FFN}^l$ denote the transformed inputs for the MHA and FFN blocks, based on equations (1-3,11), we have
\begin{equation}
{\bf{\tilde Z}}_{MHA}^l = {\rm{LN}}({{\bf{Z}}^{l - 1}}) + {\rm{LN}}({{\bf{Z}}^{l - 1}})\sum\limits_{h = 1}^p {{{\bf{D}}_h}{\bf{Q}}_h^{2l - 1}} {{\bf{U}}_h},
\label{eq12}
\end{equation}
\begin{equation}
{\bf{Z}}_{FFN}^l = {\rm{LN}}({{\bf{\tilde Z}}^l}) + {\rm{LN}}({{\bf{\tilde Z}}^l})\sum\limits_{h = 1}^p {{{\bf{D}}_h}{\bf{Q}}_h^{2l}} {{\bf{U}}_h}.
\label{eq13}
\end{equation}

Finally, the mathematical expressions of the MHA and FFN blocks become
\begin{equation}
{{\bf{\tilde Z}}^l} = {\rm{MHA}}({\bf{\tilde Z}}_{MHA}^l) + {{\bf{Z}}^{l - 1}},
\label{eq14}
\end{equation}
\begin{equation}
{{\bf{Z}}^l} = {\rm{FFN}}({\bf{Z}}_{FFN}^l) + {{\bf{\tilde Z}}^l}.
\label{eq15}
\end{equation}

During the fine-tuning stage, we freeze the parameters of the original backbone and only update the parameters contained in LRMs and the prediction head using the data of each downstream task. In the above way, the total number of introduced extra parameters is $(2dr + m{r^2})p + c$, where $c$ represents the number of parameters in the prediction head.  In LoRA, the number of extra trainable parameters is $2drm + c$. Since $r \ll d$ and $p < m$, we have $(2dr + m{r^2})p + c \approx 2drp + c < 2drm + c$. Although $rank(\sum\limits_{h = 1}^p {{{\bf{D}}_h}{\bf{Q}}_h^j} {{\bf{U}}_h}) \le pr < mr$, CLoRA strikes a balance between parameter-efficient learning and the expansion of the rank upper bound.

During the inference and storage processes, the low-rank matrices learned by CLoRA can be seamlessly merged into the corresponding weights in MHA and FFN blocks, thus avoiding extra costs. In MHA blocks, the inputs are projected into three spaces via different projection weights to obtain the query, key, and value matrices. When the $j$-th LRM of CLoRA is inserted before an MHA block, we have
\begin{equation}
\begin{array}{l}
{{\bf{X}}_{q/k/v}} = ({\bf{X}}_{input}^j + {\bf{X}}_{input}^j\sum\limits_{h = 1}^p {{{\bf{D}}_h}} {\bf{Q}}_h^j{{\bf{U}}_h}){{\bf{W}}_{q/k/v}}\\
 \qquad \quad={\bf{X}}_{input}^j({\bf{I}} + \sum\limits_{h = 1}^p {{{\bf{D}}_h}} {\bf{Q}}_h^j{{\bf{U}}_h}){{\bf{W}}_{q/k/v}},
\end{array}
\label{eq16}
\end{equation}
where ${{\bf{X}}_{q/k/v}}$ is the acquired query/key/value matrix and ${{\bf{W}}_{q/k/v}}$ denote the related projection weight. In FFN blocks, the inputs are first mapped into a high-dimensional space and then reprojected into the original space. Suppose ${{\bf{W}}_1}$ is the first projection weight and ${{\bf{X}}_{FFN,1}}$ is the representation in the high-dimensional space. When the $j$-th LRM of CLoRA is inserted before an FFN block, we have
\begin{equation}
\begin{array}{l}
{{\bf{X}}_{FFN,1}} = ({\bf{X}}_{input}^j + {\bf{X}}_{input}^j\sum\limits_{h = 1}^p {{{\bf{D}}_h}} {\bf{Q}}_h^j{{\bf{U}}_h}){{\bf{W}}_1}\\
  \qquad \quad= {\bf{X}}_{input}^j({\bf{I}} + \sum\limits_{h = 1}^p {{{\bf{D}}_h}} {\bf{Q}}_h^j{{\bf{U}}_h}){{\bf{W}}_1}.
\end{array}
\label{eq17}
\end{equation}

According to equations (16-17), the extra matrices introduced by CLoRA can be incorporated into the related projection weights contained in the pretrained ViT backbone. In other words, CLoRA doesn't change the structure of the original backbone by substituting ${{\bf{W}}_{q/k/v}}$ and ${{\bf{W}}_1}$ with $({\bf{I}} + \sum\limits_{h = 1}^p {{{\bf{D}}_h}} {\bf{Q}}_h^j{{\bf{U}}_h}){{\bf{W}}_{q/k/v}}$ and $({\bf{I}} + \sum\limits_{h = 1}^p {{{\bf{D}}_h}} {\bf{Q}}_h^j{{\bf{U}}_h}){{\bf{W}}_1}$ respectively, adding no extra computational or storage costs.

\subsection{Sample-Agnostic Diversity Enhancement }

As illustrated in equation (11), for the $j$-th LRM, its input representation update is captured by projection matrices $\{ {{\bf{D}}_h}{\bf{Q}}_h^j{{\bf{U}}_h}\} _{h = 1}^p$. The learning scheme can be regarded as a simplified form of MoE \cite{b47}, where different projection matrices represent different experts. The performance of MoE is usually degraded by the redundant information among experts, and improving expert diversity is an effective way to address this issue \cite{b52,b53}. Therefore, the focus of this section is to design diverse projection matrices.

For convenience, we let ${\bf{M}}_h^j = {{\bf{D}}_h}{\bf{Q}}_h^j{{\bf{U}}_h}$ and denote ${\bf{X}}_{input}^j$ as $({\bf{x}}_1^j; \cdots ;{\bf{x}}_a^j; \cdots ;{\bf{x}}_{n + 1}^j)$ where ${\bf{x}}_a^j \in {R^{1 \times d}}$ represents the $a$-th token. When ${\bf{x}}_a^j$ is input into the $j$-th LRM, its representations obtained by these experts are $\{ {\bf{x}}_a^j{\bf{M}}_h^j\} _{h = 1}^p$. Much redundant information will exist in these representations if they are highly similar. In other words, lower similarity implies higher diversity. In practice, the similarities among vectors are normally measured by cosine similarity. As for the $h$-th and the $r$-th representations, the similarity $s_{h,r}^{a,j}$ between them can be calculated as
\begin{equation}
s_{h,r}^{a,j} = \frac{{{\bf{x}}_a^j{\bf{M}}_h^j{{({\bf{x}}_a^j{\bf{M}}_r^j)}^{\rm{T}}}}}{{\left| {{\bf{x}}_a^j{\bf{M}}_h^j} \right|\left| {{\bf{x}}_a^j{\bf{M}}_r^j} \right|}} = \frac{{{\bf{x}}_a^j{\bf{M}}_h^j{{({\bf{M}}_r^j)}^{\rm{T}}}{{({\bf{x}}_a^j)}^{\rm{T}}}}}{{\left| {{\bf{x}}_a^j{\bf{M}}_h^j} \right|\left| {{\bf{x}}_a^j{\bf{M}}_r^j} \right|}},
\label{eq18}
\end{equation}
where ${( \cdot )^{\rm{T}}}$ denote the transpose operator and $\left|  \cdot  \right|$ represent the magnitude of a vector.

When the batch size is 1, the diversity of projection matrices $\{ {{\bf{D}}_h}{\bf{Q}}_h^j{{\bf{U}}_h}\} _{h = 1}^p$ can be naturally enhanced by minimizing the similarity regularization term $S{R^j}$ as follows.
\begin{equation}
S{R^j} = \sum\limits_{a = 1}^{n + 1} {\sum\limits_{h = 1}^p {\sum\limits_{r = h + 1}^p {{{(s_{h,r}^{a,j})}^2}} } }.
\label{eq19}
\end{equation}

When calculating $S{R^j}$, the main computational complexity is \begin{math}\mathcal{O}\end{math}$(({p^2}n + {p^2} + pn + p){d^2})$. Since this term is sample-dependent, the total computational complexity caused by a LRM is \begin{math}\mathcal{O}\end{math}$(({p^2}n + {p^2} + pn + p)b{d^2})$, where $b$ represents the batch size in the training process. Clearly, computing the similarity regularization term becomes computationally expensive as $b$ increases.

Observing equation (18), we find that ${\bf{M}}_h^j{({\bf{M}}_r^j)^{\rm{T}}}$ is the dominant factor affecting the final similarity and  ${\bf{M}}_h^j{({\bf{M}}_r^j)^{\rm{T}}} \to {\bf{0}}$ indicates $s_{h,r}^{a,j} \to 0$. Based on these observations, we propose a sample-agnostic diversity enhancement (SADE) scheme to reduce the high computational complexity induced by sample-dependent similarity. In SADE, we just focus on minimizing the distance between ${\bf{M}}_h^j{({\bf{M}}_r^j)^{\rm{T}}}$ and ${\bf{0}}$, without considering any token contained in ${\bf{X}}_{input}^j$ and any input samples. Following equation (19), we give the reformulated similarity-regularized (RSR) term as follows.
\begin{equation}
RS{R^j} = \sum\limits_{h = 1}^p {\sum\limits_{r = h + 1}^p {\left\| {{\bf{M}}_h^j{{({\bf{M}}_r^j)}^{\rm{T}}}} \right\|_F^2} } ,
\label{eq20}
\end{equation}
where $\left\|  \cdot  \right\|_F^2$ represents the squared Frobenius norm.

According to equation (20), the proposed SADE scheme needs about \begin{math}\mathcal{O}\end{math}$((0.5{p^2} + 0.5p){d^3})$. When $b > \frac{d}{{2(n + 1)}}$, $({p^2}n + {p^2} + pn + p)b{d^2} > (0.5{p^2} + 0.5p){d^3}$, meaning that SADE is more computationally efficient than the sample-dependent similarity presented in equation (19). In ViT backbones, $\frac{d}{{2(n + 1)}}$ is small enough. Taking the widely used ViT-Base, ViT-Large, and ViT-Huge backbones in \cite{b1} as examples, we report the reduced computational complexity of SADE in terms of GFLOPs in TABLE.~\ref{tab1}, compared to the sample-dependent similarity.

\begin{table}[htp]
  \centering
  \caption{Reduced GFLOPs achieved by SADE for Varying $b$.}
 \label{tab1}
    \renewcommand{\arraystretch}{1.2}
  \setlength\tabcolsep{3.0pt}
\footnotesize

    \begin{tabular}{l|c|cccccc}
    \toprule
    \textbf{Backbone}  &$\frac{d}{{2(n + 1)}}$ & $b = 2$ &$b = 4$ &$b = 8$ &$b = 16$ &$b = 32$  &$b = 64$ \\
    \hline

    ViT-Base  &1.95  &$\downarrow$2.5\%  &$\downarrow$51.3\%  &$\downarrow$75.6\%  &$\downarrow$87.8\%  &$\downarrow$93.9\% &$\downarrow$97.0\% \\
    ViT-Large &2.60  &-                   &$\downarrow$35.0\% &$\downarrow$67.5\%  &$\downarrow$83.8\%  &$\downarrow$91.9\% &$\downarrow$95.9\% \\
    ViT-Huge  &3.25  &-                   &$\downarrow$18.8\% &$\downarrow$59.4\%  &$\downarrow$79.7\%  &$\downarrow$89.9\% &$\downarrow$94.9\% \\

\bottomrule
\end{tabular}%

\end{table}%

\subsection{Objective Function}

To expand the rank upper bound and keep the parameter-efficient property, all down/up-projection weights of inserted LRMs share the same base down/up-projection matrices in CLoRA. The additional parameters brought by base-space sharing are trained with the original loss function of each downstream task, without introducing any extra loss term. In SADE, we introduce the similarity regularization term shown in equation (20) to improve the quality of the learned input representation updates. This term is optimized as part of the final loss function. Therefore, the final objective function is written as
\begin{equation}
o = los{s_{original}} + \frac{\alpha }{{{d^2}}}\sum\limits_{j = 1}^{2L} {RS{R^j}} ,
\label{eq21}
\end{equation}
where $\alpha $ is a balance parameter; $los{s_{original}}$ represents the loss of the related downstream task. For example, $los{s_{original}}$ can be the average cross-entropy loss over training samples for classification tasks.

\subsection{Comparison to Existing Work}

Some learning strategies of CLoRA share similarities with those in ARC \cite{b17} and PEGO \cite{b19}. For instance, similar to ARC, the LRMs in CLoRA are designed to learn the input representation updates of MHA and FFN blocks; the LRMs in both CLoRA and PEGO contain multiple down-projection and up-projection matrices. However, CLoRA is substantially different from the two methods. The key differences are summarized as follows.

In ARC, the LRMs for MHA (or FFN) blocks share the same symmetric down-projections and up-projections. Note that the projections applied in the two types of blocks are different. Additionally, low-dimensional re-scaling coefficients for LRMs are used to obtain block-specific representations.  Although the parameter-sharing strategy in ARC can compress the number of new parameters, the symmetry constraint limits the learning directions of projection matrices. In PEGO, the LRMs generate the final projection matrices by combining multiple independent low-rank matrices, leading to a significant increase in the number of trainable parameters. Suppose ${{\bf{G}}_f}$ and ${{\bf{G}}_v}$ are the $f$-th and the $v$-th projection matrices generated in the $j$-th LRMs. PEGO enhances the diversity between the two projection matrices by encouraging ${({{\bf{G}}_f})^{\rm{T}}}{{\bf{G}}_v}$ to approximate ${\bf{0}}$.

Following a learning scheme similar to that of LoRA, ARC learns the projection matrix $\Delta {{\bf{W}}_j}$ in the $j$-th LRM using low-rank matrix ${{\bf{A}}_j}{{\bf{B}}_j}$. However, high performance is often accompanied by a large rank upper bound, which leads to an excessive number of trainable parameters. Unlike ARC, CLoRA learns $\Delta {{\bf{W}}_j}$ through utilizing the information existing in $\{ {{\bf{A}}_i}{{\bf{B}}_i}\} _{i = 1}^m$. In addition, CLoRA significantly reduces the number of trainable parameters by the base-space sharing component. After incorporating $\{ {{\bf{A}}_i}{{\bf{B}}_i}\} _{i = 1}^m$ into all tuning modules, each LRM in CLoRA consists of a set of low-rank projection matrices. The LRMs in PEGO are trained independently, whereas in CLoRA they are jointly trained from the shared down/up-projection spaces. Similar to PEGO, CLoRA adopts the SADE component to improve the diversity of the learned information in each LRM. But the adopted diversity-induced strategies in PEGO and CLoRA are fundamentally different. For simplicity, we also use the above generated projection matrices ${{\bf{G}}_f}$ and ${{\bf{G}}_v}$ to exemplify the strategy of SADE. Under this strategy, we expect to minimize the distance between ${{\bf{G}}_f}{({{\bf{G}}_v})^{\rm{T}}}$ and ${\bf{0}}$ to decrease the similarity between the representations extracted by ${{\bf{G}}_f}$ and ${{\bf{G}}_v}$. However, PEGO just focuses on the orthogonality between the columns of ${{\bf{G}}_f}$ and ${{\bf{G}}_v}$.

\section{Experiments}

Extensive experiments are carried out to validate the learning performance of our proposed method in this section. Since ViT is originally designed for visual classification tasks, existing studies \cite{b17,b21,b38,b39} typically evaluate their methods on VTAB-1K benchmark \cite{b54} and fine-grained visual classification (FGVC) benchmark \cite{b12}. For fair comparisons, we also evaluate CLoRA on the two benchmarks. As introduced in Section IV, CLoRA includes base-space sharing and SADE components. Ablation studies are conducted to further analyze our proposed method. Unless otherwise specified, we use the ViT-B/16 backbone \cite{b1} pre-trained on ImageNet-21K dataset for 2D visual tasks. Recently, ViT has been extended to the field of point cloud analysis. Since the network structures of these extensions are similar to the original ViT, CLoRA can be directly incorporated into their pre-trained backbones. Following IDPT \cite{b55}, DAPT \cite{b41} and PointGST \cite{b56}, we also test CLoRA's fine-tuning performance on the tasks of point cloud object recognition and part segmentation. For comparison methods, we directly adopt their reported results when available; otherwise, we run their released code following the parameter settings provided in the original papers.

\subsection{Experiments on the VTAB-1K Benchmark}

VTAB-1K is composed of 19 visual classification datasets which are categorized into three groups: \textbf{Natural} (containing 7 datasets), \textbf{Specialized} (containing 4 datasets) and \textbf{Structured} (containing 8 datasets). More details are provided in TABLE.~\ref{tab2}, where ``\#Class", ``\#Training", ``\#Validation" and ``\#Test" represent the number of classes, training samples, validation samples and test samples, respectively. We follow the same standard data augmentations as described in VPT \cite{b12} to maintain consistency.

\begin{table}[!htb]
  \centering
  \caption{Details of the VTAB-1K Benchmark.}
  \label{tab2}
    \renewcommand{\arraystretch}{1.2}
  \setlength\tabcolsep{3pt}
\footnotesize
  \begin{tabular}{lccccc}
    \toprule
    \textbf{Dataset}      &\textbf{Group}   &\#\textbf{Class}   &\#\textbf{Training}  &\#\textbf{Validation}   &\#\textbf{Test}  \\ \hline

   Cifar100    &\multirow{7}{*}{Natural}     &100      &\multirow{7}{*}{800/1000}     &\multirow{7}{*}{200}  &10000     \\

   Caltech101    &                            &102     &  &        &6084     \\

   DTD          &                             &47      &   &         &1880    \\
   Flower102    &                             &102     &  &           &6149    \\

    Pets       &                             &37       &   &        & 3669   \\
    SVHN       &                             &10       &  &         &26032    \\
    Sun397    &                              &397      &  &          &21750    \\
       \hline

  Camelyon         &\multirow{4}{*}{Specialized}      &2     &\multirow{4}{*}{800/1000}   &\multirow{4}{*}{200}   &32768 \\

   EuroSAT           &           &10        &     &    &5400 \\
   Resisc45          &           &45        &     &    &6300 \\
   Retinopathy       &           &5        &     &    &42670 \\
      \hline
  Clevr-Count         &\multirow{8}{*}{Structured}            &8         &\multirow{8}{*}{800/1000}   &\multirow{8}{*}{200}     &15000     \\
  Clevr-Dist         &           &6        &     &   &15000  \\
  DMLab              &           &6        &     &    &22735 \\
  KITTI-Dist         &           &4        &     &    &711 \\
  dSpr-Loc           &           &16        &     &    &73728 \\
  dSpr-Ori           &           &16        &     &    &73728 \\
  sNORB-Azim         &           &18        &     &    &12150 \\
  sNORB-Ele          &           &9        &     &    &12150 \\

\bottomrule
  \end{tabular}
\end{table}

\begin{table}[!htb]
  \centering
  \caption{Parameter settings of CLoRA on 2D visual tasks.}
  \label{tab3}
\footnotesize
    \renewcommand{\arraystretch}{1.2}
  \setlength\tabcolsep{2pt}
  \begin{tabular}{lc}
    \toprule
    \textbf{Name}      &\textbf{Value}  \\ \hline

    Optimizer          &Adam                           \\
Learning Rate          &\{0.03,0.01,0.005,0.003,0.001\}\\
Weight Decay           &\{0.05,0.01,0.005,0.001\}\\
Batch Size             &32  \\
Learning Rate Schedule	&Cosine Decay\\
Warmup Epochs	&10\\
Training Epochs	&100\\
$\alpha $	&\{0.1,1.0,10.0\}\\
$p$	 &\{4,6,8,10\}\\
$r$	 &8\\

\bottomrule
  \end{tabular}
\end{table}

Fully fine-tuning the whole pre-trained backbone (denoted as Full) and tuning the linear prediction head only (denoted as Linear) are two traditional fine-tuning methods. Here, we compare CLoRA with the two traditional fine-tuning methods and competitive PEFT baselines. The selected PEFT methods are LoRA \cite{b13}, VPT \cite{b12}, ARC \cite{b17}, NOAH \cite{b42}, RLRR \cite{b39}, SCT \cite{b40}, SA2VP \cite{b21}, DoRA \cite{b14}, PEGO \cite{b19}, LoRA+AOFT \cite{b45} and CDRA-SPT \cite{b38}. In TABLE.~\ref{tab3}, we give the parameter settings of CLoRA. Following VPT, we first adopt the training (800)-validation (200) data split to determine the suitable parameter settings, then conduct tuning experiments on the entire 1000 training data using these settings and finally report the top-1 test accuracy on each dataset.

%c|ccccccc|cccc|cccccccc|ccc
\begin{table*}[!htb]
  \centering
  \caption{Experimental results on the VTAB-1K Benchmark. ``Mean" represents the average accuracy of the 19 datasets. ``Param. (M)" denotes the number of trainable parameters in millions and ``\#Param. (M)" means the average Param. (M). ``GFLOPs" represents the average GFLOPs across all datasets. In parameter-efficient tuning methods, \textbf{bold font} and \underline{underline} denote the \textbf{best} and the \underline{second-best} results. }
  \label{tab4}
  \renewcommand{\arraystretch}{1.2}
  \setlength\tabcolsep{3pt}
\footnotesize
  \begin{tabular}{l|ccccccc|cccc|cccccccc|cc}
    \toprule
    &\multicolumn{7}{c|}{{\textbf{Natural}}}  &\multicolumn{4}{c|}{{\textbf{Specialized}}}  &\multicolumn{8}{c|}{{\textbf{Structured}}}
    &

    \\
       \textbf{Method}        &\rotatebox{90}{Cifar100}   &\rotatebox{90}{Caltech101}     &\rotatebox{90}{DTD}    &\rotatebox{90}{Flower102}   &\rotatebox{90}{Pets}   &\rotatebox{90}{SVHN}   &\rotatebox{90}{Sun397} &\rotatebox{90}{Camelyon}&\rotatebox{90}{EuroSAT}&\rotatebox{90}{Resisc45}&\rotatebox{90}{Retinopathy}&\rotatebox{90}{Clevr-Count}&\rotatebox{90}{Clevr-Dist}&\rotatebox{90}{DMLab}&\rotatebox{90}{KITTI-Dist}&\rotatebox{90}{dSpr-Loc}&\rotatebox{90}{dSpr-Ori}&\rotatebox{90}{sNORB-Azim}&\rotatebox{90}{sNORB-Ele}&\rotatebox{90}{\textbf{Mean}(\%)}&\rotatebox{90}{\textbf{\#Param. (M)}}\\

              \hline
     \multicolumn{22}{c}{\textit{Traditional fine-tuning methods}}   \\\hline

    Full  &68.9 &87.7 &64.3 &97.2 &86.9 &87.4 &38.8 &79.7 &95.7 &84.2 &73.9 &56.3 &58.6 &41.7 &65.5 &57.5 &46.7 &25.7 &29.1  &65.6 &85.8
      \\
    Linear &63.4 &85.0 &63.2 &97.0 &86.3 &36.6 &51.0 &78.5 &87.5 &68.6 &74.0 &34.3 &30.6 &33.2 &55.4 &12.5 &20.0 &9.6 &19.2 &52.9 &0.04         \\\hline
    \multicolumn{22}{c}{\textit{Parameter-efficient tuning methods}}   \\\hline

    LoRA \cite{b13} &67.1 &91.4 &69.4 &98.8 &90.4 &85.3 &54.0 &84.9 &95.3 &84.4 &73.6 &\underline{82.9} &\textbf{69.2} &49.8 &78.5 &75.7 &47.1 &31.0 &44.0 &72.3 &0.33
    \\

    VPT \cite{b12}  &\textbf{78.8} &90.8 &65.8 &98.0 &88.3 &78.1 &49.6  &81.8 &
\underline{96.1} &83.4 &68.4 &68.5 &60.0 &46.5 &72.8 &73.6 &47.9 &32.9 &37.8 &69.4 &0.60  \\

    ARC \cite{b17} &72.2 &90.1 &
72.7 &99.0 &91.0 &\textbf{91.9} &54.4 &84.9 &95.7 &\textbf{86.7} &75.8 &80.7 &67.1 &48.7 &81.6 &79.2 &51.0 &31.4 &39.9 &73.4 &
0.13
    \\

    NOAH \cite{b42} &69.6 &
92.7 &70.2 &99.1 &90.4 &86.1 &53.7 &84.4 &95.4 &83.9 &75.8 &82.8 &\underline{68.9} &49.9 &81.7 &81.8 &48.3 &32.8 &44.2 &73.3 &0.50
    \\
   RLRR \cite{b39}&75.6 &92.4 &72.9 &\underline{99.3} &
91.5 &89.8 &\underline{57.0} &
86.8 &95.2 &85.3 &75.9 &79.7 &64.2 &\textbf{53.9} &\textbf{82.1} &
83.9 &\textbf{53.7} &33.4 &43.6 &\underline{74.5} &0.33
   \\
   SCT \cite{b40} &75.3 &91.6 &72.2 &99.2 &91.1 &91.2 &55.0 &85.0 &
\underline{96.1} &86.3 &
\underline{76.2} &81.5 &65.1 &
51.7 &80.2 &75.4 &46.2 &33.2 &
\textbf{45.7} &73.6 &0.15
   \\

SA2VP \cite{b21} &73.0 &91.9 &70.5 &99.1 &90.8 &84.7 &56.8 &86.0 &95.9 &85.8 &75.2 &76.6 &61.8 &50.8 &79.9 &\underline{84.5} &52.8 &\textbf{34.7} &\underline{45.3} &73.5&0.48 \\

   DoRA \cite{b14} &66.1 &93.1 &68.8 &97.0 &89.9 &87.2 &56.4 &83.1 &94.5 &80.8 &75.2 &79.1 &62.1 &48.0 &80.6 &83.1 &51.8 &33.2 &44.0 &72.3&0.19 \\
PEGO \cite{b19} &75.1 &92.9 &\underline{73.1} &\textbf{99.4} &\underline{92.1} &89.2 &56.5 &\underline{87.0} &\textbf{96.2} &86.1 &76.1 &78.4 &62.4 &52.7 &79.5 &80.6 &52.3 &31.9 &37.7 &73.6 &0.44
\\
   LoRA+AOFT \cite{b45} &73.6 &92.6 &71.1 &\underline{99.3} &91.3 &85.3 &56.9  &84.7 &94.8 &83.7 &75.6 &76.7 &63.2 &48.7 &81.0 &82.2 &\underline{53.1} &26.9 &45.1 &72.9&\textbf{0.08} \\

   CDRA-SPT \cite{b38} &73.8 &\underline{93.3} &72.7 &\textbf{99.4} &91.6 &89.9 &55.8   &86.8 &\textbf{96.2} &86.2 &76.0  &\textbf{83.0} &\underline{68.9} &51.5 &\textbf{82.1} &80.9 &51.7 &33.0 &43.2 &\underline{74.5}&0.49 \\

         \rowcolor{gray!20}
 CLoRA &\underline{76.3}&\textbf{93.8}&\textbf{73.4}&\textbf{99.4}&\textbf{92.4}&\underline{91.3} &\textbf{57.8} &\textbf{87.4}&\textbf{96.2}&\underline{86.4}&\textbf{76.4} &81.6&67.3&\underline{52.9}&\underline{81.9}&\textbf{84.7}&\textbf{53.7}&\underline{33.9}&40.1&\textbf{75.1} &
\underline{0.11}
   \\

\bottomrule
  \end{tabular}
\end{table*}

The comparison results on the VTAB-1K benchmark are reported in TABLE.~\ref{tab4}. From these experimental results, we have the following observations. 1) PEFT methods have obvious advantages over full fine-tuning (Full) and linear probing (Linear). Although full fine-tuning introduces a large number of trainable parameters, it tends to be negatively affected by overfitting due to the lack of sufficient training data. Linear probing reduces the number of trainable parameters by only tuning the linear prediction head, but it cannot obtain task-specific representations. These drawbacks explain why both linear probing and full fine-tuning obtain the worst and the second-worst mean results. 2) CLoRA achieves much more competitive classification results than all comparison methods. Specifically, it exceeds the two second-best methods (RLRR and CDRA-SPT) by 0.6\% in terms of mean accuracy while further reducing the number of trainable parameters by 66.67\% and 77.55\%, respectively. Moreover, CLoRA obtains either the highest or second-highest accuracy on 16 out of 19 datasets. 3) LoRA+AOFT and CLoRA have the smallest and the second-smallest numbers of trainable parameters. However, compared with LoRA+AOFT, CLoRA achieves a 2.2\% improvement in terms of mean accuracy. Note that CLoRA, ARC, RLRR, DoRA, PEGO and LoRA+AOFT are LoRA-based methods. CLoRA clearly performs better than LoRA and its variations (excluding LoRA+AOFT) in terms of both mean accuracy and parameter efficiency.

\subsection{Experiments on the FGVC Benchmark}

FGVC benchmark consists of five fine-grained visual classification datasets including CUB-200-2011, NABirds, Oxford Flowers, Stanford Dogs and Stanford Cars. TABLE.~\ref{tab5} shows the details of these datasets. Since the effectiveness of LoRA, VPT, ARC, RLRR, SA2VP, LoRA+AOFT and CDRA-SPT has been verified on the five datasets, they are selected as comparison methods. For the purpose of comparison, samples from the datasets are processed by the same data augmentations used in VPT.

\begin{table}[!htb]
  \centering
  \caption{Details of the FGVC Benchmark.}
  \label{tab5}
 \footnotesize
      \renewcommand{\arraystretch}{1.2}
  \setlength\tabcolsep{3pt}
  \begin{tabular}{lcccc}
    \toprule
    \textbf{Dataset}         &\#\textbf{Class}   &\#\textbf{Training}  &\#\textbf{Validation}   &\#\textbf{Test}  \\ \hline

   CUB-200-2011 (CUB)      &200     &5394    &600  &5794    \\

   NABirds (NBirds)        &555     &21536  &2393        &24633    \\
   Oxford Flowers (Flowers)   &102      &1020  &1020 &6149\\
   Stanford Dogs (Dogs)        &120     &10800      &1200  &8580\\
   Stanford Cars (Cars)         &196      &7329   &815  &8041\\

\bottomrule
  \end{tabular}
\end{table}

We present the experimental results of comparison methods and CLoRA in TABLE.~\ref{tab6}.  CLoRA outperforms the second-best method (RLRR) by 0.4\% in terms of mean accuracy while reducing the number of trainable parameters by 46.81\%. In addition, CLoRA achieves the highest classification accuracies on Flowers and Dogs datasets and obtains the second-best classification accuracies on the other datasets.

\begin{table}[t]
  \centering
  \caption{Experimental results on the FGVC benchmark.}
 \label{tab6}
    \renewcommand{\arraystretch}{1.2}
  \setlength\tabcolsep{2.2pt}
\footnotesize

    \begin{tabular}{l|ccccc|cc}
    \toprule
    \textbf{Method}  & CUB & NBirds & Flowers & Dogs & Cars & \textbf{Mean}(\%)  & \textbf{Param.(M)} \\
    \hline

     \multicolumn{8}{c}{\textit{Traditional fine-tuning methods}}   \\\hline
    Full  & 87.3  & 82.7  & 98.8  & 89.4  & \underline{84.5}  & 88.5 & 85.98 \\
    Linear  & 85.3  & 75.9  & 97.9  & 86.2  & 51.3  & 79.3 & 0.18 \\
       \hline

  \multicolumn{8}{c}{\textit{Parameter-efficient tuning methods}}   \\\hline
    LoRA  & 88.3 & 85.6  & 99.2   & 91.0  & 83.2  & 89.5 & 0.44 \\
    VPT  & 88.5 & 84.2  & 99.0    & 90.2  & 83.6  & 89.1 & 0.85 \\

    ARC  & 88.5  & 85.3  & 99.3  & 91.9  & 85.7  & 90.1 & \underline{0.25} \\
   RLRR & \textbf{89.3}  & 84.7  & \textbf{99.5}  & \underline{92.0}  & 87.0  & \underline{90.4} & 0.47 \\

   SA2VP & 89.1  & \textbf{85.8}  & 99.3  & \textbf{92.1}  & 84.1  & 90.1 & 0.85 \\

    LoRA+AOFT & 88.8  & 84.2  &\underline{99.4}  & \underline{92.0}  & 85.1  & 89.9  & \textbf{0.22} \\

    CDRA-SPT   & 88.6  & 83.4 & 99.3  & 89.8  & \textbf{88.2}  & 89.9 & 0.74 \\

    \rowcolor{gray!20}
     CLoRA  & \underline{89.2}  & \underline{85.7}  & \textbf{99.5}  & \textbf{92.1}  & \underline{87.3}  & \textbf{90.8}&\underline{0.25}  \\

\bottomrule
\end{tabular}%

\end{table}%

\subsection{Experiments on Larger-Scale ViT Backbones}

We further validate the performance of CLoRA on larger-scale ViT backbones. LoRA, VPT, ARC, RLRR, and LoRA+AOFT have been extensively validated on ViT-Large and ViT-Huge backbones, showing strong performance on the VTAB-1K benchmark. For comparison consistency, we also conduct experiments on the same benchmark with the two larger ViT backbones. Comparison results on ViT-Large and ViT-Huge backbones are reported in TABLE.~\ref{tab7} and TABLE.~\ref{tab8}. From the experimental results, we observe that CLoRA achieves much more competitive classification performance than the comparison methods and simultaneously introduces the smallest number of trainable parameters.

\begin{table}[!htb]
  \centering
  \caption{Comparison results on the ViT-Large backbone. }
  \label{tab7}
 \footnotesize
     \renewcommand{\arraystretch}{1.2}
  \setlength\tabcolsep{1.2pt}
  \begin{tabular}{l|ccc|cc}

    \toprule
    \textbf{Method}    &Natural(7)   &Specialized(4)  &Structured(8)   &\textbf{Mean}(\%) &\textbf{\#Param.(M)}  \\ \hline
     \multicolumn{6}{c}{\textit{Traditional fine-tuning methods}}   \\\hline

   Full     &74.7  &83.8      &48.1    &65.4 &303.40		  \\

   Linear       &70.9  &69.1      &25.8    &51.5 &0.05		  \\\hline
\multicolumn{6}{c}{\textit{Parameter-efficient tuning methods}}   \\\hline
   LoRA     &70.5  &85.0      &57.3    &72.0 &0.74		  \\
   VPT      &82.5  &83.9      &54.1    &70.8 &0.49			  \\
   ARC       &82.3  &85.6      &57.3    &72.5 &\underline{0.18}	  \\
   RLRR        &\underline{83.9}  &\underline{86.4}      &\textbf{61.9}    &\underline{75.2} &0.82		  \\
   LoRA+AOFT &83.3  &85.9      &60.2    &74.3 &\textbf{0.15}	  \\
   \rowcolor{gray!20}
   CLoRA &\textbf{84.7}  &\textbf{86.6}      &\underline{61.5}    &\textbf{75.3} &\textbf{0.15} \\
\bottomrule
  \end{tabular}
\end{table}

\begin{table}[!htb]
  \centering
  \caption{Comparison results on the ViT-Huge backbone. }
  \label{tab8}
\footnotesize
     \renewcommand{\arraystretch}{1.2}
  \setlength\tabcolsep{1.2pt}
  \begin{tabular}{l|ccc|cc}

    \toprule
    \textbf{Method}    &Natural(7)   &Specialized(4)  &Structured(8)   &\textbf{Mean}(\%) &\textbf{\#Param.(M)}  \\ \hline

    \multicolumn{6}{c}{\textit{Traditional fine-tuning methods}}   \\\hline
   Full     &70.9  &83.6      &46.0    &63.1 &630.90		  \\

   Linear       &67.9  & 79.0     &26.1    &52.7 &0.06		  \\\hline
\multicolumn{6}{c}{\textit{Parameter-efficient tuning methods}}   \\\hline

   LoRA         &77.1  &83.5      &55.4    &69.3 &1.21	 \\
   VPT      &77.9  &83.3      &52.2    &68.2 &0.96			  \\
   ARC       &79.1  &84.8&53.7    &69.9 &0.22	  \\
   RLRR        &\underline{79.4}  &\underline{85.1}      &\textbf{59.0}    &\underline{72.0} &1.33		  \\
   LoRA+AOFT &78.8  &83.8      &58.3   &71.3 &\underline{0.20}		  \\
   \rowcolor{gray!20}
      CLoRA &\textbf{80.7}  &\textbf{85.6}      &\underline{58.4}    &\textbf{72.3} &\textbf{0.19} \\

\bottomrule
  \end{tabular}
\end{table}

\subsection{Ablation Studies}

CLoRA includes base-space sharing and SADE components. In base-space sharing, two LRMs are inserted before the MHA and FFN blocks in each encoder layer, and all down/up-projection weights in LRMs share the same down/up-projection spaces. To improve the diversity of the information learned from the shared spaces, SADE is introduced in the training process. Here, taking the VTAB-1K benchmark as an example, we first conduct ablation experiments to investigate the influence of different parts and then verify the effectiveness of base-space sharing and SADE components.

The results of ablation experiments are reported in TABLE.~\ref{tab9}. Note that notations are listed in the first column for convenience. ``$\checkmark$'' means the corresponding item is selected, while ``$\times$'' indicates it is not selected.  For instance, in CLoRAMF, two LRMs are inserted before MHA and FFN blocks while SADE is not applied. According to the results, we observe that CLoRA exceeds CLoRAMF, CLoRAMS and CLoRAFS by 1.7\%, 2.8\% and 2.6\%, respectively. This observation implies that adapting the input representations of FFN blocks is more important than the others and SADE has the least influence.

\begin{table}[!htb]
  \centering
  \caption{Ablation experiments on the VTAB-1K benchmark.}
  \label{tab9}
 \footnotesize
     \renewcommand{\arraystretch}{1.2}
  \setlength\tabcolsep{2.2pt}
  \begin{tabular}{l|ccc|c}

    \toprule
      \textbf{Notation}  &Inserted Before MHA  &Inserted Before FFN  &SADE   &\textbf{Mean}(\%)\\ \hline

   CLoRAMF &$\checkmark$ &$\checkmark$ &$\times$       &73.4  \\

    CLoRAMS    &$\checkmark$  &$\times$  &$\checkmark$ &72.3     \\
     CLoRAFS  &$\times$     &$\checkmark$    &$\checkmark$    	&72.5\\
        \rowcolor{gray!20}
    CLoRA  &$\checkmark$    &$\checkmark$ &$\checkmark$ &\textbf{75.1}\\

\bottomrule
  \end{tabular}
\end{table}

Based on LoRA, each LRM in PEGO combines a group of low-rank projection matrices which are regularized by column orthogonality. LRMs in LoRA and its variants including PEGO are usually adopted to update the query and value projection weights. Following an information-learning scheme similar to LoRA, CLoRA can be easily applied to learn the query and value weight updates by replacing the identity matrices with the corresponding projection weights. Directly calculating the similarities among the representations extracted by multiple projection matrices is computationally expensive for relatively large batch sizes. In contrast, SADE can efficiently approximate these relationships. Here, we further evaluate the performance of the two components designed for CLoRA.

\begin{figure}[!htb]
\centering
\includegraphics[trim={0.1cm 0.50cm 0.0cm 0.36cm},clip,width=3.4in]{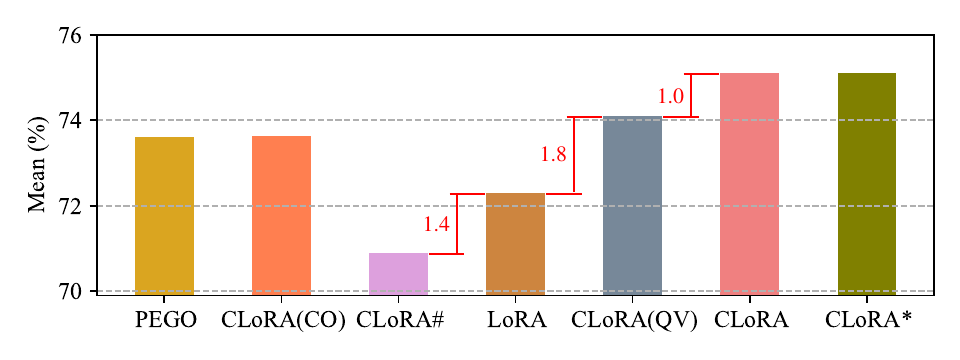}
\caption{Performance verification of the two components in CLoRA.}
\label{fig3}
\end{figure}

The comparison results on the VTAB-1K benchmark in terms of mean accuracy are shown in Fig.~\ref{fig3}. CLoRA(QV) denotes the variant where LRMs are employed to learn the query and value weight updates. CLoRA(CO) replaces the SADE component with the column orthogonality regularization used in PEGO. Additionally, CLoRA(CO) follows the learning scheme provided by PEGO. By comparing PEGO, CLoRA(CO), CLoRA(QV) and LoRA, we observe that: 1) CLoRA(CO) achieves similar performance to PEGO; 2) CLoRA(QV) evidently outperforms PEGO, CLoRA(CO) and LoRA. These comparisons illustrate the effectiveness of the base-space sharing and SADE components. From the results of CLoRA (QV) and CLoRA, we observe that inserting LRMs to tune the input representations for MHA and FFN blocks enables learning more informative representations than those learned through the query and value weight updates. CLoRA\# represents that the final low-rank matrices obtained by LRMs are directly learned by $\{ {{\bf{A}}_i}{{\bf{B}}_i}\} _{i = 1}^m$ (see equation (7)). According to these results, CLoRA exceeds CLoRA\# by 4.2\% and CLoRA\# even underperforms LoRA, which illustrates that our designed base-space sharing component can address the potential optimization conflicts effectively. CLoRA* means that the similarities are directly calculated using equation (18). Compared with CLoRA*, CLoRA performs comparably while achieving a 93.9\% reduction in GFLOPs (see TABLE.~\ref{tab1}), indicating the efficiency brought by SADE.

\begin{table*}[tp]

  \centering
  \renewcommand{\arraystretch}{1.2}
  \setlength\tabcolsep{3.1pt}
\footnotesize
  \caption{Classification accuracy (\%) on ModelNet40 dataset and three variants of ScanObjectNN dataset. \#Param. (M) and \#GFLOPs denote the average number of trainable parameters and the average GFLOPs across all datasets.}
    \label{tab10}
    \begin{tabular}{lccccccccc}
    \toprule
    \multirow{2}{*}{Pre-trained model} &\multirow{2}{*}{Fine-tuning strategy} &\multirow{2}{*}{Reference} &\multirow{2}{*}{\#Param. (M)} &\multirow{2}{*}{\#GFLOPs} &\multicolumn{3}{c}{ScanObjectNN} &\multirow{2}{*}{ModelNet40}\\
    \cmidrule(lr){6-8}
    & & & & &OBJ\_BG & OBJ\_ONLY &PB\_T50\_RS     \\
    \hline
   \multirow{7}{*}{\tabincell{c}{Point-MAE~\cite{b61}}} & \textcolor{gray}{Full}  &- & \textcolor{gray}{22.1 (100\%)}  & \textcolor{gray}{4.76} & \textcolor{gray}{90.02} & \textcolor{gray}{88.29} & \textcolor{gray}{85.18} & \textcolor{gray}{93.2}\\
    & IDPT~\cite{b55} &ICCV 23  & 1.7 (7.69\%) & 7.10($\uparrow$) &91.22\dplus{+1.20} &\underline{90.02}\dplus{+1.73} & 84.94{\dtplus{-0.24}}  &93.3\dtplus{-0.1}\\

    & Point-PEFT~\cite{b63} &AAAI 24 & 0.7 (3.13\%)  & 7.61($\uparrow$) & 89.67{\dtplus{-0.35}} & 88.98\dplus{+0.69} & 84.91{\dtplus{-0.27}} &93.3\dplus{+0.1} \\
    & DAPT~\cite{b41} &CVPR 24& 1.1 (4.97\%)  & 4.96($\uparrow$) & 90.88\dplus{+0.86} &\textbf{90.19}\dplus{+1.90} & 85.08{\dtplus{-0.10}}  & \underline{93.5}{\dplus{+0.3}}  \\
    & PointGST~\cite{b56}  &TPAMI 25&\underline{0.6} (2.77\%) &\underline{4.81}($\uparrow$) &\textbf{91.74}\dplus{+1.72} &\textbf{90.19}\dplus{+1.90} & 85.29\dplus{+0.11}  & \underline{93.5}{\dplus{+0.3}} \\

    & PointLoRA~\cite{b64} &CVPR 25 & 0.8 (3.43\%)  &5.06($\uparrow$) & 90.71\dplus{+0.69} &89.33\dplus{+1.04} &\textbf{85.53}\dplus{+0.35}  & 93.3{\dplus{+0.1}} \\

    & CLoRA(\textbf{ours})&- & \textbf{0.3} (1.36\%) & \textbf{4.76}(-) &\underline{91.57}\dplus{+1.55} & \textbf{90.19}\dplus{+1.90} &\underline{85.32}\dplus{+0.14}  & \textbf{93.6}{\dplus{+0.4}} \\
    \hline

    \multirow{7}{*}{\tabincell{c}{Point-BERT~\cite{b60}}} & \textcolor{gray}{Full}  &- & \textcolor{gray}{22.1 (100\%)}  & \textcolor{gray}{4.76} & \textcolor{gray}{87.43} & \textcolor{gray}{88.12} & \textcolor{gray}{83.07} & \textcolor{gray}{92.7}\\
    & IDPT~\cite{b55} &ICCV 23  & 1.7 (7.69\%) & 7.10($\uparrow$) & 88.12\dplus{+0.69} & 88.30\dplus{+0.18} & 83.69\dplus{+0.62}   & 92.6{\dtplus{-0.1}}\\
    & Point-PEFT~\cite{b63} &AAAI 24 & 0.7 (3.13\%)  & 7.61($\uparrow$) & 88.81\dplus{+1.38} &\underline{89.67}\dplus{+1.55} & 85.00\dplus{+1.93} &\textbf{93.4}\dplus{+0.7} \\
    & DAPT~\cite{b41} &CVPR 24& 1.1 (4.97\%)  & 4.96($\uparrow$) & 91.05\dplus{+3.62} &\underline{89.67}\dplus{+1.55} & 85.43\dplus{+2.36}  & 93.1{\dplus{+0.4}}  \\
    & PointGST~\cite{b56}  &TPAMI 25&\underline{0.6} (2.77\%)  &\underline{4.81}($\uparrow$) &\underline{91.39}\dplus{+3.96} &\underline{89.67}\dplus{+1.55} & \underline{85.64}\dplus{+2.57}  & \textbf{93.4}{\dplus{+0.7}} \\

    & PointLoRA~\cite{b64} &CVPR 25 & 0.9 (4.07\%)  & 5.06($\uparrow$) & 89.85\dplus{+2.42} &88.98\dplus{+0.86} &84.63\dplus{+1.56}  &\underline{93.2}{\dplus{+0.5}} \\

    & CLoRA(\textbf{ours}) &- &\textbf{0.3} (1.36\%)  & \textbf{4.76}(-) & \textbf{91.74}\dplus{+4.31} & \textbf{90.36}\dplus{+2.24} & \textbf{85.80}\dplus{+2.73}  & \textbf{93.4}{\dplus{+0.7}} \\

    \hline

    \multirow{7}{*}{\tabincell{c}{RECON~\cite{b62}}} & \textcolor{gray}{Full}  &- & \textcolor{gray}{22.1 (100\%)}  & \textcolor{gray}{4.76} & \textcolor{gray}{94.32} & \textcolor{gray}{92.77} & \textcolor{gray}{90.01} & \textcolor{gray}{92.5}\\
    & IDPT~\cite{b55} &ICCV 23  & 1.7 (7.69\%) & 7.10($\uparrow$) & 93.29{\dtplus{-1.03}} & 91.57{\dtplus{-1.20}} & 87.27{\dtplus{-2.74}}   & 93.4\dplus{+0.9}\\

    & Point-PEFT~\cite{b63} &AAAI 24 & 0.7 (3.13\%)  & 7.61($\uparrow$) & 91.91{\dtplus{-2.41}} & 90.19{\dtplus{-2.58}} & 86.36{\dtplus{-3.65}} &93.3\dplus{+0.8} \\
    & DAPT~\cite{b41} &CVPR 24& 1.1 (4.97\%)  & 4.96($\uparrow$) &\underline{94.32}\dplus{+0.00} & 92.43{\dtplus{-0.34}} & 89.38{\dtplus{-0.63}}  & 93.5{\dplus{+1.0}}  \\
    & PointGST~\cite{b56}  &TPAMI 25&\underline{0.6} (2.77\%)  &\underline{4.81}($\uparrow$) &\textbf{94.49}\dplus{+0.17} &\textbf{92.94}\dplus{+0.17} & \underline{89.49}{\dtplus{-0.52}}  &\textbf{93.6}{\dplus{+1.1}} \\

    & PointLoRA~\cite{b64} &CVPR 25 & 0.8 (3.43\%)  &5.06($\uparrow$)  &93.46{\dtplus{-0.86}}  &91.22{\dtplus{-1.55}}  &88.65{\dtplus{-1.36}}   &93.4\dplus{+0.9}  \\

    & CLoRA(\textbf{ours}) &- & \textbf{0.4} (1.81\%)  & \textbf{4.76}(-) & \underline{94.32}\dplus{+0.00} & \underline{92.77}\dplus{+0.00} & \textbf{89.51}{\dtplus{-0.50}}  & \textbf{93.8}{\dplus{+1.3}} \\

\bottomrule
    \end{tabular}
\end{table*}

\begin{table}[!htb]
  \centering
  \caption{Parameter settings of CLoRA on point cloud datasets.}
  \label{tab11}
\footnotesize
  \renewcommand{\arraystretch}{1.2}
  \setlength\tabcolsep{2pt}
  \begin{tabular}{lc}
    \toprule
    \textbf{Name}      &\textbf{Value}  \\ \hline
\multirow{2}{*}{\tabincell{c}{Number of Points}}     &1024 (ModelNet40)\\
                        &2048 (ScanObjectNN\&ShapeNetPart)\\\hhline{|~|-|}
\multirow{2}{*}{\tabincell{c}{Batch Size}} &32 (ModelNet40\&ScanObjectNN)\\

             &16 (ShapeNetPart) \\\hhline{|~|-|}
\multirow{2}{*}{\tabincell{c}{Learning Rate}} &0.0005 (ModelNet40\&ScanObjectNN)\\

         &0.0002 (ShapeNetPart)\\\hhline{|~|-|}

Weight Decay           &0.05 \\
Optimizer          &AdamW                           \\
Learning Rate Schedule	&Cosine Decay\\
Warmup Epochs	&10\\
Training Epochs	&300\\
$\alpha $	&\{0.01,0.1,1.0,10.0\}\\
$p$	 &\{6,8,10,12\}\\
$r$	 &8\\

\bottomrule
  \end{tabular}
\end{table}

\subsection{Further Exploration}

Since pre-trained point cloud transformers share similar encoder blocks with ViT, CLoRA can be directly applied to adapt these models for downstream tasks. Consistent with related methods \cite{b41,b55}, we verify the performance of CLoRA on three widely used point cloud datasets: ModelNet40 \cite{b57}, ScanObjectNN \cite{b58} and ShapeNetPart \cite{b59}. Both ModelNet40 and ScanObjectNN are point cloud object recognition datasets. ModelNet40 consists of 12311 CAD models from 40 classes. All point clouds in ModelNet40 are complete, uniform and noise-free. Besides, they are independently and identically distributed. ScanObjectNN contains approximately 15000 real-world point cloud instances covering 15 indoor object categories. It includes three variants (OBJ\_BG, OBJ\_ONLY, PB\_T50\_RS) arranged in order of increasing difficulty. ShapeNetPart is a widely used dataset for point-level part segmentation and it comprises 16881 samples across 16 object types and 50 part categories.

Following IDPT \cite{b55} and DAPT \cite{b41}, we select Point-BERT \cite{b60}, Point-MAE \cite{b61}, and RECON \cite{b62} as pre-trained backbones. IDPT, Point-PEFT \cite{b63}, DAPT, PointGST \cite{b56} and PointLoRA \cite{b64} are excellent PEFT methods for point cloud learning and are therefore selected as comparison methods in this subsection. Additionally, full fine-tuning (Full) serves as the baseline. For fair comparisons, we adopt the same data augmentation techniques used in \cite{b41}. Note that all experimental results are reported without voting strategies.

\noindent\textit{1) Experiments on Point Cloud Object Recognition}

The classification results of the comparison methods and CLoRA are presented in TABLE.~\ref{tab10}. Based on IDPT and DAPT, the parameter settings of CLoRA used in these experiments are provided in TABLE.~\ref{tab11}. All methods are compared using three metrics: 1) the average number of trainable parameters (\#Param.(M)); 2) the average GFLOPs (\#GFLOPs); and 3) top-1 classification accuracy. Observing these results, we find that CLoRA achieves the highest or the second-highest classification accuracy across different pre-trained backbones and datasets. Among the comparison methods, PointGST introduces the fewest trainable parameters when fine-tuning Point-BERT, Point-MAE and RECON backbones. Nevertheless, compared with PointGST, CLoRA further reduces the parameter count by 50\%, 50\%, and 33.3\%, respectively, demonstrating its strong parameter efficiency. Although IDPT, Point-PEFT, DAPT, PointGST and PointLoRA perform well in terms of classification accuracy and the number of trainable parameters, they bring extra computational cost at the inference stage. In contrast, the trainable modules designed for CLoRA can be seamlessly incorporated into the original backbones at the inference stage, without introducing extra cost.

We further evaluate the few-shot learning ability of our proposed method by conducting few-shot experiments on the ModelNet40 dataset. Following the same data settings provided by Zha et al. \cite{b54}, we report the learning results in TABLE.~\ref{tab12}. In low-data scenarios, CLoRA achieves the highest accuracy in most cases across the three backbones when compared to IDPT, Point-PEFT, DAPT, and PointLoRA. While PointGST performs better than CLoRA on the Point-MAE backbone, CLoRA outperforms PointGST on the RECON backbone. These results demonstrate the effectiveness of CLoRA under low-data regimes.

\begin{table}[!htp]

  \centering
 \footnotesize
  \renewcommand{\arraystretch}{1.2}
  \setlength\tabcolsep{3.6pt}
  \caption{Few-shot learning results on the ModelNet40 dataset. Results are presented as mean classification accuracy (\%)$\pm$standard deviation (\%) over 10 independent experiments.}
    \label{tab12}
  % \vspace{-10pt}
    \begin{tabular}{lccccc}
    \toprule
   \multirow{2}{*}{Method}&\multirow{2}{*}{Reference} & \multicolumn{2}{c}{5-way} & \multicolumn{2}{c}{10-way} \\
\cmidrule(lr){3-4}\cmidrule(lr){5-6}  &        & 10-shot & 20-shot & 10-shot & 20-shot \\
    \hline
    \multicolumn{6}{c}{\textit{Pre-trained Point-MAE backbone}} \\
        \textcolor{gray}{+ Full}&\textcolor{gray}{ECCV 22} & \textcolor{gray}{96.3$\pm$2.5} & \textcolor{gray}{97.8$\pm$1.8} & \textcolor{gray}{92.6$\pm$4.1} & \textcolor{gray}{95.0$\pm$3.0}\\
   + IDPT &   ICCV 23    & \underline{97.3}$\pm$2.1& 97.9$\pm$1.1&92.8$\pm$4.1& 95.4$\pm$2.9\\
   + Point-PEFT  & AAAI 24 & 95.5$\pm$2.9 & 97.6$\pm$1.7 & 91.7$\pm$4.3 & 94.7$\pm$3.0 \\
   + DAPT & CVPR 24 & 96.8$\pm$1.8  & 98.0$\pm$1.0 &\underline{93.0}$\pm$3.5 & 95.5$\pm$3.2  \\
   + PointGST & TPAMI 25 &\textbf{98.0}$\pm$1.8 &\textbf{98.3}$\pm$0.9 &\textbf{93.7}$\pm$4.0 &\textbf{95.7}$\pm$2.4 \\
   + PointLoRA &CVPR 25 &96.6$\pm$2.6  &97.7$\pm$1.3 &92.0$\pm$4.2 &95.2$\pm$3.3\\
      \rowcolor{gray!20}
   + CLoRA  &-  &96.8$\pm$1.9 &\underline{98.1}$\pm$1.1   &92.2$\pm$4.1  &\underline{95.6}$\pm$2.9 \\

    \hline

       \multicolumn{6}{c}{\textit{Pre-trained Point-BERT backbone}} \\
   \textcolor{gray}{+ Full}  &\textcolor{gray}{CVPR 22} &\textcolor{gray}{94.6$\pm$3.1} & \textcolor{gray}{96.3$\pm$2.7} & \textcolor{gray}{91.0$\pm$5.4} & \textcolor{gray}{92.7$\pm$5.1} \\
   + IDPT  & ICCV 23    & 96.0$\pm$1.7& 97.2$\pm$2.6& 91.9$\pm$4.4& 93.6$\pm$3.5\\
   + Point-PEFT  & AAAI 24 & 95.4$\pm$3.0 & 97.3$\pm$1.9 & 91.6$\pm$4.5 & 94.5$\pm$3.5 \\
   + DAPT & CVPR 24 &95.8$\pm$2.1 &97.3$\pm$1.3&92.2$\pm$4.3&94.2$\pm$3.4 \\
   + PointGST  &TPAMI 25&\textbf{96.5}$\pm$2.4 &\underline{97.9}$\pm$2.0 &\textbf{92.7}$\pm$4.2 &\underline{95.0}$\pm$2.8 \\
   + PointLoRA &CVPR 25 &95.7$\pm$2.5  &97.8$\pm$1.7  &92.0$\pm$4.1 &94.7$\pm$3.0\\
      \rowcolor{gray!20}
     + CLoRA  &-  &\underline{96.1}$\pm$2.3 &\textbf{98.0}$\pm$1.7  &\underline{92.4}$\pm$4.0  &\textbf{95.1}$\pm$2.7\\
    \hline

          \multicolumn{6}{c}{\textit{Pre-trained RECON backbone}} \\
       \textcolor{gray}{+ Full}  &\textcolor{gray}{ICML 23} &\textcolor{gray}{93.7$\pm$1.9} & \textcolor{gray}{98.9$\pm$1.2} & \textcolor{gray}{93.3$\pm$3.9} & \textcolor{gray}{95.8$\pm$3.0} \\
   + IDPT  & ICCV 23    & 96.0$\pm$1.7& 97.2$\pm$2.6& 91.9$\pm$4.4& 93.6$\pm$3.5\\
   + Point-PEFT  & AAAI 24 & 95.4$\pm$3.0 & 97.3$\pm$1.9 & 91.6$\pm$4.5 & 94.5$\pm$3.5 \\
   + DAPT & CVPR 24 &95.8$\pm$2.1 &97.3$\pm$1.3&92.2$\pm$4.3&94.2$\pm$3.4 \\
   + PointGST  &TPAMI 25&96.5$\pm$2.4 &97.9$\pm$2.0 &\underline{92.7}$\pm$4.2 &95.0$\pm$2.8 \\
   + PointLoRA &CVPR 25 &\textbf{96.9}$\pm$2.7 &\textbf{98.8}$\pm$1.2 &\underline{92.7}$\pm$4.4 &\textbf{95.8}$\pm$2.9\\
   \rowcolor{gray!20}
     + CLoRA  &-   &\underline{96.7}$\pm$2.1  &\underline{98.2}$\pm$1.5  &\textbf{92.8}$\pm$4.1 &\underline{95.7}$\pm$3.0\\
    \bottomrule
    \end{tabular}

\end{table}

\noindent \textit{2) Experiments on Point Cloud Part Segmentation}

TABLE~\ref{tab13} presents the part segmentation results on the ShapeNetPart dataset, following the parameter settings provided in TABLE~\ref{tab11}.  Full fine-tuning almost outperforms all PEFT methods including CLoRA. However, it requires a large number of trainable parameters, which is inefficient for adapting pre-trained backbones to downstream tasks and imposes significant storage overhead. Therefore, we focus on comparing the learning performance of PEFT methods. According to the experimental results, CLoRA achieves comparable or superior performance compared to the best among IDPT, Point-PEFT, DAPT, PointGST, and PointLoRA. More importantly, CLoRA has an obvious advantage in reducing the number of trainable parameters.

\begin{table}[htp]

  \centering
 \footnotesize
  \renewcommand{\arraystretch}{1.2}
    \setlength\tabcolsep{2.8pt}
  \caption{Part segmentation results on the ShapeNetPart dataset. Cls. mIoU (\%) and Inst. mIoU (\%) represent the class-averaged and instance-averaged mIoU, respectively. }
  \label{tab13}
    \begin{tabular}{lcccc}
    \toprule
    Method & Reference & Params. (M)& Cls. mIoU (\%) & Inst. mIoU (\%) \\
    \hline
       \multicolumn{5}{c}{\textit{Pre-trained Point-MAE backbone}} \\
   \textcolor{gray}{+ Full} &  \textcolor{gray}{ECCV 22} & \textcolor{gray}{27.06} & \textcolor{gray}{84.19} & \textcolor{gray}{86.1} \\
    + IDPT & ICCV 23 & 5.69  & 83.79  & \underline{85.7}  \\
    + Point-PEFT & AAAI 24 & 5.62  & 83.20 &  85.2 \\
    + DAPT & CVPR 24 & 5.65  &\underline{84.01} & \underline{85.7} \\
    + PointGST & TPAMI 25 &\underline{5.59}  & 83.81 & \textbf{85.8} \\
    + PointLoRA &CVPR 25 &5.71 &83.74 &85.3\\
       \rowcolor{gray!20}
    + CLoRA &-  &\textbf{5.29} &\textbf{84.03} &\underline{85.7}\\
    \hline

\multicolumn{5}{c}{\textit{Pre-trained Point-BERT backbone}} \\
    \textcolor{gray}{+ Full} &  \textcolor{gray}{CVPR 22} & \textcolor{gray}{27.06} & \textcolor{gray}{84.11} & \textcolor{gray}{85.6} \\
    + IDPT & ICCV 23 & 5.69  & 83.50  & 85.3  \\
    + Point-PEFT & AAAI 24 & 5.62  & 81.12  &  84.3 \\
    + DAPT & CVPR 24 & 5.65  &\underline{83.83} &\underline{85.5} \\
    + PointGST &TPAMI 25 & \underline{5.58}  & \textbf{83.87} & \textbf{85.7} \\
    + PointLoRA &CVPR 25 &5.68 &82.61&85.2\\
       \rowcolor{gray!20}
    + CLoRA &- &\textbf{5.30}&83.81 &\underline{85.5} \\

      \hline

\multicolumn{5}{c}{\textit{Pre-trained RECON backbone}} \\
    \textcolor{gray}{+ Full} &  \textcolor{gray}{ICML 23} & \textcolor{gray}{27.06} & \textcolor{gray}{84.52} & \textcolor{gray}{86.1} \\
    + IDPT & ICCV 23 & 5.69  & 83.66  & \underline{85.7}  \\
    + Point-PEFT & AAAI 24 & 5.62  & 83.10  &  85.1 \\
    + DAPT & CVPR 24 & 5.65  & \underline{83.87} &\underline{85.7} \\
    + PointGST &TPAMI 25 &\underline{5.58}  &\textbf{83.98} & \textbf{85.8} \\
    + PointLoRA &CVPR 25  &5.63  &\textbf{83.98} &85.4 \\
       \rowcolor{gray!20}
    + CLoRA &- &\textbf{5.29}&\textbf{83.98}&85.6 \\
    \bottomrule
    \end{tabular}

\end{table}

\section{Conclusion and future work}

A novel parameter-efficient fine-tuning method termed collaborative low-rank adaptation (CLoRA) is proposed in this paper. CLoRA includes two key components: base-space sharing and SADE. In base-space sharing, all LRMs share the same down/up-projection spaces to increase their learning capacities and reduce the number of trainable parameters. Each LRM in CLoRA constructs multiple low-rank matrices from the shared spaces and combines the representations extracted by these matrices. To further improve the performance of LRMs, SADE is applied to regularize the relationship among the combined representations. Note that all learned low-rank matrices of CLoRA can be seamlessly incorporated into the related weights of fine-tuned backbones, without introducing extra inference and storage costs. Experimental results on image and point cloud benchmarks illustrate that CLoRA has an excellent ability to keep a balance between learning performance and parameter efficiency. Additionally, CLoRA achieves the state-of-the-art inference efficiency on point cloud benchmarks.

Similar to existing PEFT methods, CLoRA performs well on downstream tasks when provided with high-quality data. Nevertheless, in some practical application scenarios, such as underwater target recognition and autonomous driving, the collected samples are often of low quality, posing significant challenges for effective fine-tuning. In future work, we will focus on designing suitable representation-learning strategies to extract robust task-specific representations by leveraging the knowledge contained in pre-trained models.

% or



\ifCLASSOPTIONcaptionsoff
  \newpage
\fi

\end{document}